\documentclass[letterpaper,journal]{IEEEtran}
\usepackage{amsmath,amsfonts}
\usepackage{algorithmic}
\usepackage{algorithm}
\usepackage{array}
\usepackage{textcomp}
\usepackage{stfloats}
\usepackage{url}
\usepackage{verbatim}
\usepackage{graphicx}
\usepackage{cite}
\usepackage{balance}
\usepackage{multicol}
\usepackage{xcolor}
\usepackage{multirow}
\usepackage{subfigure}
\usepackage[colorlinks=true, citecolor=blue, linkcolor=blue, urlcolor=blue]{hyperref}
\usepackage[all]{hypcap}
\usepackage{array}
\usepackage{makecell}
\usepackage{tabularray}
\usepackage{graphicx}
\usepackage{xcolor}
\usepackage{eso-pic}

\newcommand{\YMBioPhoto}[1]{\includegraphics[width=1in,height=1.25in,clip]{#1}}

\makeatletter
\let\oldhypertarget\hypertarget
\renewcommand{\hypertarget}[2]{\def\Hy@temp{#2}\ifx\Hy@temp\@empty
    \Hy@raisedlink{\oldhypertarget{#1}{}}\else
    \oldhypertarget{#1}{#2}\fi
}

\def\YM@currentbibkey{}\let\YM@orig@bibitem\@bibitem
\renewcommand{\@bibitem}[1]{\def\YM@currentbibkey{#1}\YM@orig@bibitem{#1}}
\renewcommand{\@biblabel}[1]{\ifx\YM@currentbibkey\@empty
    [#1]\else
    [\hyperlink{citeback:\YM@currentbibkey}{#1}]\fi
}
\makeatother

\AddToShipoutPictureFG{%
  \AtPageUpperLeft{%
    \put(\LenToUnit{0.75in},\LenToUnit{-0.34in}){%
      \makebox(0,0)[tl]{%
        \parbox{0.78\paperwidth}{%
          \raggedright
          \normalsize
          $>$ This work has been submitted to the IEEE for possible publication.
          Copyright may be transferred without notice, after which this version
          may no longer be accessible.%
        }%
      }%
    }%
  }%
}
\begin{document}

\title{Multi-Solution Inverse Design of Pixelated Patch Antennas Using an Autoregressive Transformer}

\author{Young-Mok~Yun, Seung-Won~Shin,
Jaeyong~Chung,~\IEEEmembership{Member,~IEEE},
and Kyung-Sik~Choi,~\IEEEmembership{Member,~IEEE}\thanks{This work was in part supported by the Institute of Information \& Communications
Technology Planning \& Evaluation (IITP) grant funded by the Korea government
(MSIT) under Grant RS-2025-09942968, AI Semiconductor Innovation
Lab (Yonsei University), in part by the National Research Foundation of Korea (NRF) funded by the Ministry of Science and ICT (MSIT), Korea Government, under Grant RS-2025-00560232. Jaeyong Chung and Kyung-Sik Choi are co-corresponding authors.}\thanks{Young-Mok~Yun, Seung-Won~Shin, Jaeyong~Chung, and Kyung-Sik~Choi are with the Department of System Semiconductor Engineering, Yonsei University, Seoul 03722, South Korea.}}

\markboth{}{}

\maketitle

\begin{abstract}
Deep-learning-based inverse design has emerged as a promising framework for pixelated patch antenna synthesis by enabling broader exploration of discrete geometry spaces and reducing design time compared with conventional template-based methods. Existing methods, including surrogate-assisted metaheuristic optimization, tandem inverse networks, and diffusion models, have advanced the field but still face practical limitations. Metaheuristic methods such as binary particle swarm optimization (BPSO) are sensitive to initialization and search hyperparameters and are often time-consuming, tandem networks have difficulty capturing the one-to-many nature of inverse antenna design, and diffusion models may require excessive complexity and data for moderate-scale binary layout synthesis. In this paper, an autoregressive (AR) transformer is presented for the inverse design of $\mathbf{10 \times 10}$ pixelated patch antennas. Its sequential discrete generation process enables the synthesis of multiple feasible structures for a given target response. The pretrained model is further fine-tuned for dual-band design using step-shaped target responses with sharp spectral transitions, allowing intuitive band specification without manually constructing detailed $S_{11}$ curves. Compared with BPSO, our proposed method generates valid and diverse antenna candidates more efficiently. Fabricated and measured examples confirm that distinct pixelated layouts can realize the desired antenna responses, thereby demonstrating the multi-solution capability of the proposed framework.
\end{abstract}

\begin{IEEEkeywords}
Antennas, generative AI, design automation, microstrip antennas, inverse design, multifrequency antennas.
\end{IEEEkeywords}

\section{Introduction}

\IEEEPARstart{I}{n} recent years, machine learning has played an increasingly important role in the design of electromagnetic (EM) and photonic structures, offering new possibilities beyond conventional simulation-driven and template-based methodologies\hypertarget{citeback:karahan2024deep}{}\cite{karahan2024deep},\hypertarget{citeback:zhou2026ai}{}\cite{zhou2026ai},\hypertarget{citeback:zhou202525}{}\cite{zhou202525},\hypertarget{citeback:yuan2025tandem}{}\cite{yuan2025tandem},\hypertarget{citeback:karahan2023deep}{}\cite{karahan2023deep},\hypertarget{citeback:karahan2022deep}{}\cite{karahan2022deep},\hypertarget{citeback:karahan2024deep_2}{}\cite{karahan2024deep_2},\hypertarget{citeback:svelto2025ai}{}\cite{svelto2025ai},\hypertarget{citeback:chu2025ai}{}\cite{chu2025ai},\hypertarget{citeback:chae2024pulserf}{}\cite{chae2024pulserf},\hypertarget{citeback:chae2025ml}{}\cite{chae2025ml}. In antenna engineering, conventional synthesis typically relies on repeated electromagnetic simulations and substantial domain knowledge to tune geometry parameters, making it difficult to scale to large design spaces and complex target specifications. This challenge becomes particularly critical in multiband antenna design, where multiple impedance-matching notches must be precisely aligned with the desired operating frequencies while ensuring sufficient bandwidth and practical radiation characteristics. To overcome the limited flexibility of predefined templates, pixelated antenna design has been explored as a high-degree-of-freedom approach by representing the radiating structure as a binary pixel array\hypertarget{citeback:ullah2023multiservice}{}\cite{ullah2023multiservice},\hypertarget{citeback:chen2026surrogate}{}\cite{chen2026surrogate},\hypertarget{citeback:wang2024machine}{}\cite{wang2024machine},\hypertarget{citeback:gjokaj2017design}{}\cite{gjokaj2017design}. However, this flexibility introduces a large discrete combinatorial design space, making direct trial-and-error or purely simulation-driven optimization inefficient. To address this challenge, deep-learning-based inverse design has been widely investigated, including surrogate-assisted metaheuristic optimization\hypertarget{citeback:gupta2024transfer}{}\cite{gupta2024transfer},\hypertarget{citeback:chen2025inverse}{}\cite{chen2025inverse},\hypertarget{citeback:fu2022efficient}{}\cite{fu2022efficient}, tandem inverse networks\cite{yuan2025tandem},\hypertarget{citeback:gupta2023tandem}{}\cite{gupta2023tandem},\hypertarget{citeback:deng2024automated}{}\cite{deng2024automated} for one-shot geometry prediction, and diffusion-based generative models\hypertarget{citeback:kim2025diffusion}{}\cite{kim2025diffusion},\hypertarget{citeback:guo2025dall}{}\cite{guo2025dall},\hypertarget{citeback:pang2025inverse}{}\cite{pang2025inverse} capable of synthesizing multiple feasible candidates for a given target response.

\begin{figure}[t]
    \centering
    \setlength{\fboxsep}{1pt}
    \setlength{\fboxrule}{0.8pt}
    \fcolorbox{white}{white}{\includegraphics[width=0.98\linewidth]{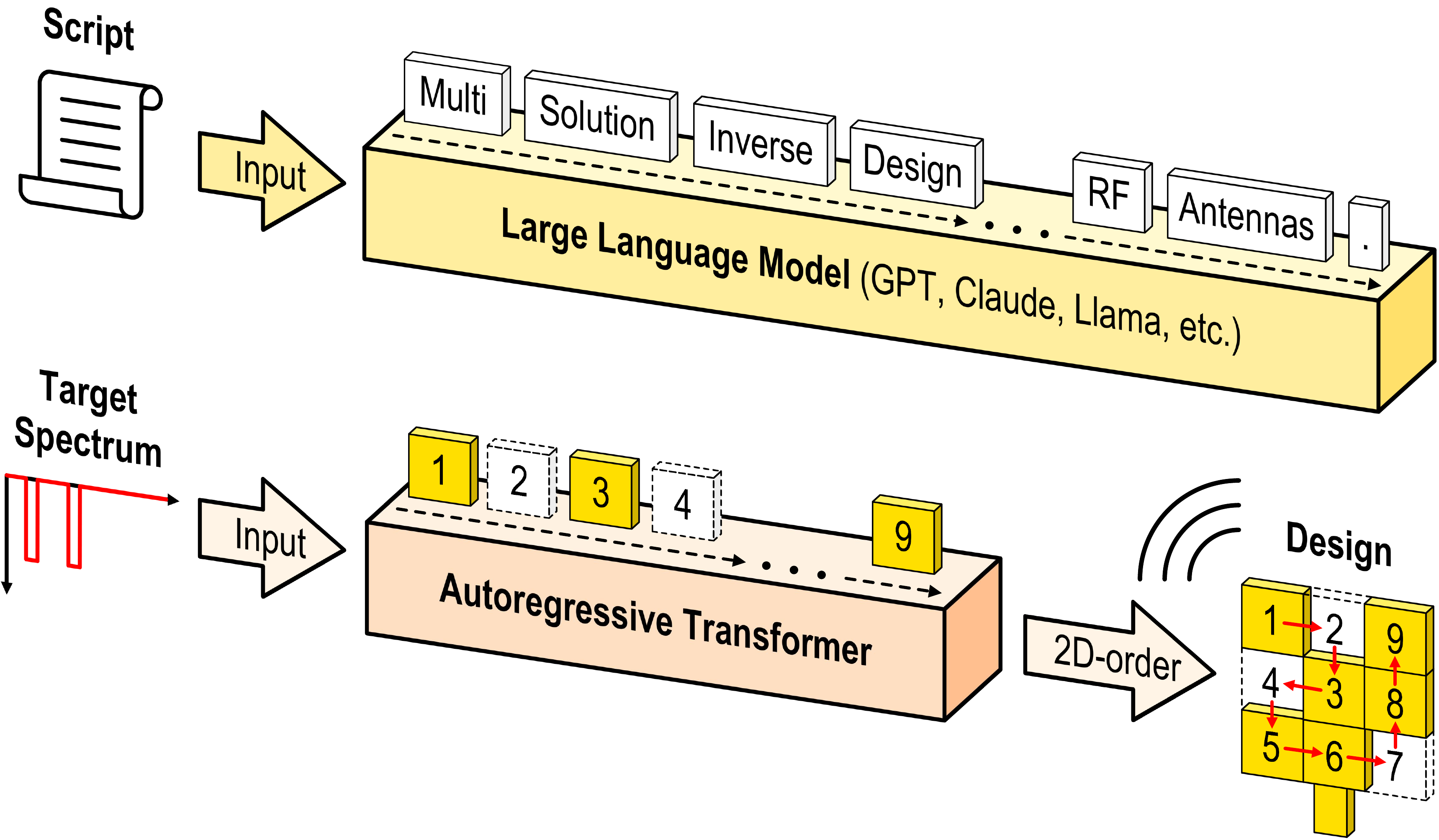}
    }
    \caption{LLM-inspired autoregressive framework for antenna inverse design.}
    \label{fig:LLM-inspired}
\end{figure}

Despite recent progress, existing inverse design approaches still have several limitations. 
Surrogate-assisted iterative optimization can be sensitive to model parameters and may become trapped in local optima\hypertarget{citeback:ghewari2025advancements}{}\cite{ghewari2025advancements},\hypertarget{citeback:kim2025inverse}{}\cite{kim2025inverse}, while tandem inverse networks typically provide only a single representative solution for a given target, limiting design diversity in inherently nonunique inverse antenna design problems. 
This limitation is critical in practical antenna design because geometries with similar $S_{11}$ responses can differ in radiation characteristics, bandwidth, gain and robustness to fabrication-induced variations. 
Thus, generating multiple feasible candidates enables designers to rapidly compare alternative layouts and select a final structure that satisfies the target frequency response while also providing favorable bandwidth, radiation performance, and fabrication robustness\hypertarget{citeback:chen2026deep}{}\cite{chen2026deep}. 
While diffusion-based models can also provide multiple candidates, they typically require large datasets and multi-stage training, which is heavier than necessary for moderate-scale pixelated layouts.

To address these issues, this work adopts an autoregressive (AR) transformer that models the conditional distribution of discrete antenna layouts under a desired target response. As illustrated in Fig.~\ref{fig:LLM-inspired}, the proposed framework is inspired by large language models, where a sequence of tokens is generated autoregressively conditioned on a given prompt; similarly, the AR transformer performs inverse antenna design through a conditional pixel-by-pixel generation scheme. In this scheme, each antenna pixel is sequentially determined based on the target reflection spectrum and the previously generated pixels, allowing the model to account for inter-pixel dependencies during geometry construction. Unlike deterministic inverse models that directly map a target response to a single layout, this probabilistic generation process learns the conditional joint distribution of the full pixelated structure and can produce multiple valid antenna geometries corresponding to the same target spectrum. Therefore, the proposed pixel generation method naturally reflects the one-to-many solution space inherent in inverse antenna design and provides multiple design alternatives for user-specified single- and dual-band responses. The model is further fine-tuned using step-shaped target responses, enabling intuitive band-level target specification without manually constructing detailed $S_{11}$ curves. In this way, the desired operating frequency bands can be specified directly, and multiple antenna candidates can be generated in a short time for subsequent selection based on both impedance matching and practical antenna performance. The proposed framework is validated through single- and dual-band antenna synthesis, comparison with binary particle swarm optimization (BPSO)-based optimization, and fabricated measurement.

The rest of this paper is organized as follows.
Section~\ref{sec:proposed_models} describes the architectures and training procedures of the proposed inverse and forward models.
Section~\ref{sec:implementation} compares the proposed AR transformer with BPSO in terms of generation efficiency and design performance.
Section~\ref{sec:measurement} reports fabrication and measurement results of generated antennas.
Section~\ref{sec:discussion} discusses model design considerations and further analysis.
Finally, Section~\ref{sec:conclusion} concludes this paper.

\section{Proposed Machine Learning Models}\label{sec:proposed_models}

\begin{figure*}[!t]
    \centering
    \includegraphics[width=\textwidth]{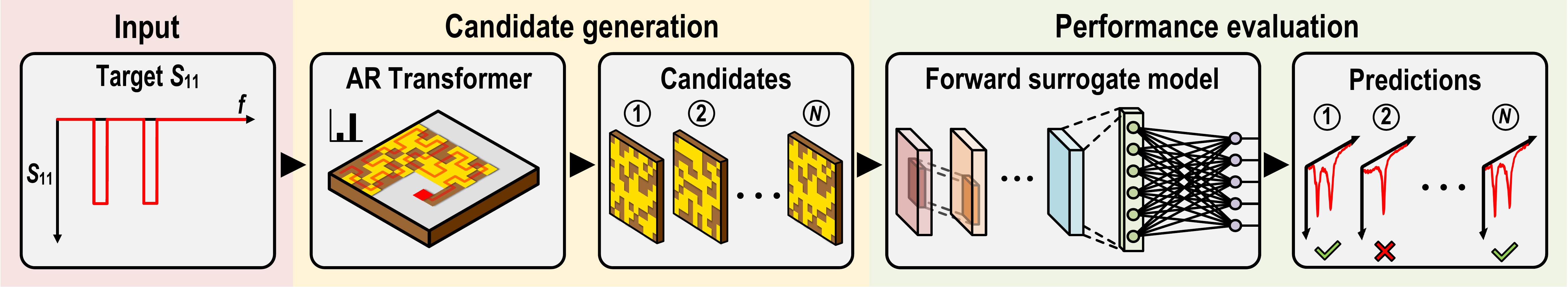}
    \caption{Overview of the proposed framework.}
    \label{fig:overview}
\end{figure*}

\begin{figure}[t]
    \centering
    \includegraphics[width=\linewidth]{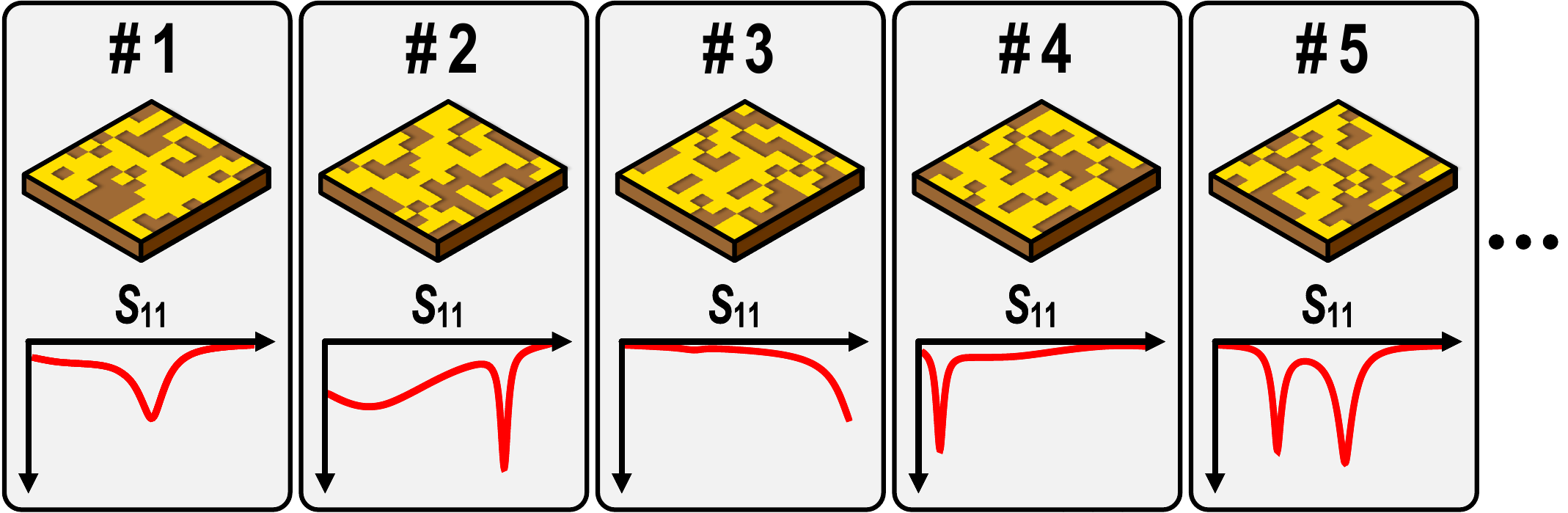}
    \caption{Pixel-spectrum paired dataset for training.}
    \label{fig:dataset}
\end{figure}

As shown in Fig.~\ref{fig:overview}, the proposed framework consists of candidate generation (inverse network) and performance evaluation (forward network). The inverse network is implemented using an AR transformer, which is well suited for the one-to-many nature of antenna inverse design because multiple pixelated geometries can produce similar reflection responses. Instead of directly mapping a target spectrum to a single deterministic layout, the AR transformer generates the antenna geometry pixel by pixel, where each pixel is conditioned on both the target reflection spectrum and the previously generated pixels. This sequential conditional generation allows the model to capture inter-pixel dependencies and learn the joint distribution of physically meaningful antenna layouts.

A step-function-like $S_{11}$ spectrum is used as the input condition to represent the desired operating bands. Based on this target response, the AR transformer generates multiple pixelated antenna candidates. These candidates are subsequently evaluated by a forward surrogate model, which rapidly predicts their reflection spectra and compares them with the target response without requiring full-wave EM simulation for every design. Finally, the candidates that best satisfy the target specification are selected as valid antenna designs.

To construct the training dataset, 500k random pixelated antenna layouts are generated, as illustrated in Fig.~\ref{fig:dataset}. 
Each generated layout is then analyzed using HFSS full-wave EM simulation to obtain the corresponding $S_{11}$ response, forming a pixel--spectrum paired dataset. Simulations are parallelized across multiple accounts on two CPU-based computing servers: an Intel Xeon Silver 4310 system at 2.10~GHz with 12 cores and 24 logical processors, and a dual-socket Intel Xeon Gold 6242R system at 3.10~GHz with 40 cores and 80 logical processors. 
This parallelized dataset generation produces approximately 10k samples per day. The dataset is divided into training, validation, and test sets with a ratio of 80\%, 10\%, and 10\%, respectively, and is used to train both the inverse and forward surrogate models. 
Detailed simulation settings for dataset generation are described in Section~\ref{sec:implementation}.

\begin{figure}[t]
    \centering
    \includegraphics[width=\linewidth]{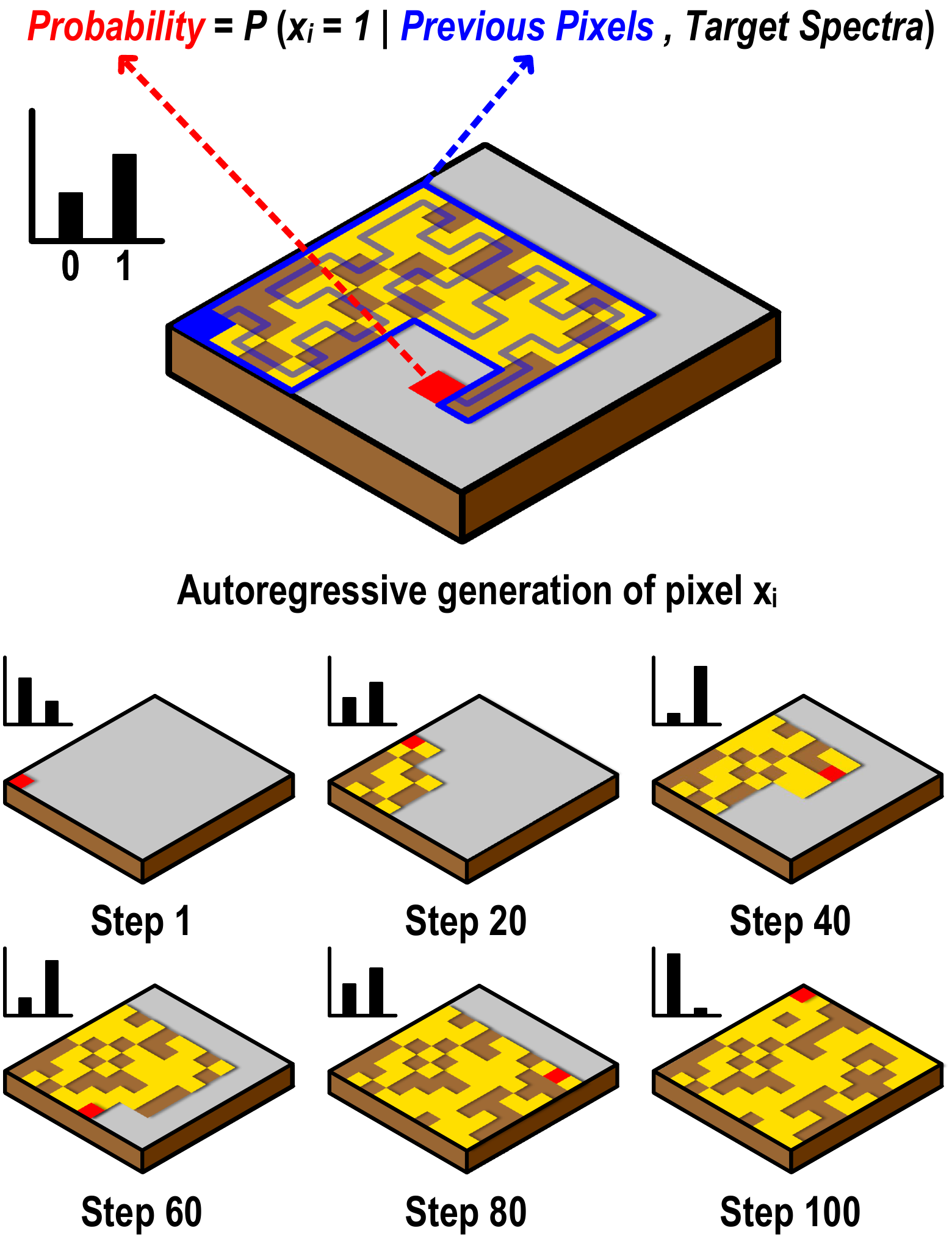}
    \caption{Illustration of the autoregressive pixel generation process.}
    \label{fig:drawing_test_main}
\end{figure}

\subsection{Inverse Model}

Fig.~\ref{fig:drawing_test_main} illustrates the sequential generation process of the proposed autoregressive transformer. Instead of predicting the entire antenna layout in a single step, the inverse model sequentially generates the antenna layout pixel by pixel. At each generation step shown in Fig.~\ref{fig:drawing_test_main}, the red pixel denotes the current prediction target, the blue region represents the previously generated pixels, and the gray region indicates the pixels that remain ungenerated. Given the target spectrum and the previously generated pixel configuration, the model determines whether the current pixel should be present or absent in the antenna layout. The selected value is then appended to the partial layout and used as additional context for the next prediction step.

\begin{figure}[t]
    \centering
    \includegraphics[width=\linewidth]{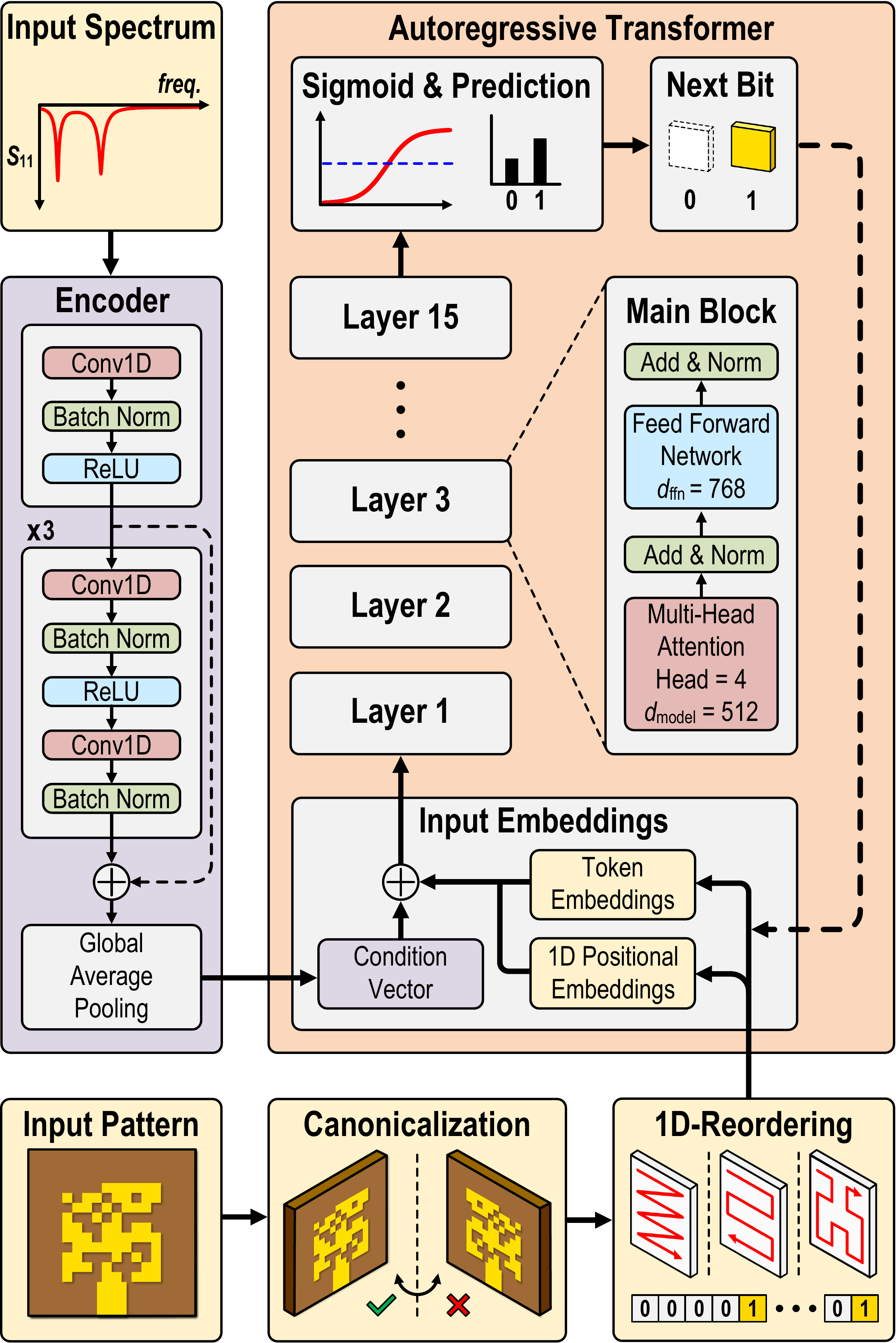}
    \caption{Detailed architecture of the AR inverse model.}
    \label{fig:final_inverse}
\end{figure}

Fig.~\ref{fig:final_inverse} illustrates the architecture of the proposed inverse model. A target input reflection coefficient ($S_{11}$) spectrum is first encoded by a 1-D ResNet-based spectral encoder into a compact condition vector. Through residual blocks and global average pooling operation, the encoder extracts spectral features that represent the desired antenna response. The encoded condition vector is then broadcast to all sequence positions and added to the token embeddings and positional embeddings. In this way, the spectral condition is injected into the transformer at every pixel-generation step, so that each next-pixel prediction is guided by both the target response and the partial pixel layout generated up to the current step.

Since the inverse model generates the antenna layout in an autoregressive manner, the two-dimensional pixel grid must be converted into an ordered one-dimensional sequence before training. This serialization defines the generation order of the pixels, allowing the model to predict each pixel based on the target spectrum and the previously generated pixels. Thus, the model learns layout generation as sequential next-pixel prediction rather than one-shot prediction of the entire pixel pattern. The $10 \times 10$ layout is represented as a one-dimensional sequence of 100 binary values, denoted as $[x_1, x_2, \dots, x_{100}]$, where each $x_i$ represents the state of the $i$-th pixel after serialization. Here, $x_i=1$ and $x_i=0$ indicate the presence and absence of a metallic patch element at the $i$-th pixel position, respectively.

Before serialization, canonicalization\hypertarget{citeback:kaba2023equivariance}{}\cite{kaba2023equivariance},\hypertarget{citeback:ma2024canonicalization}{}\cite{ma2024canonicalization} under left--right symmetry is applied to the two-dimensional pixel layout. For each binary layout, its horizontally flipped counterpart is generated. The two layouts are then converted into binary sequences, and the one that comes first in lexicographic order is selected as the canonical target. Note that this procedure differs from data augmentation. Instead, it reduces ambiguity in the target space by mapping symmetry-equivalent layouts to a single representative form. This is especially important in the early generation steps, because two left--right symmetric antenna layouts can produce identical spectral responses, but they may correspond to different binary sequences. By using canonicalization, the model receives consistent supervision and the ambiguity of the inverse mapping is greatly reduced.

\begin{figure}[t]
\centering
\setlength{\fboxsep}{1pt}
\setlength{\fboxrule}{0.8pt}
\fcolorbox{white}{white}{\includegraphics[width=0.98\linewidth]{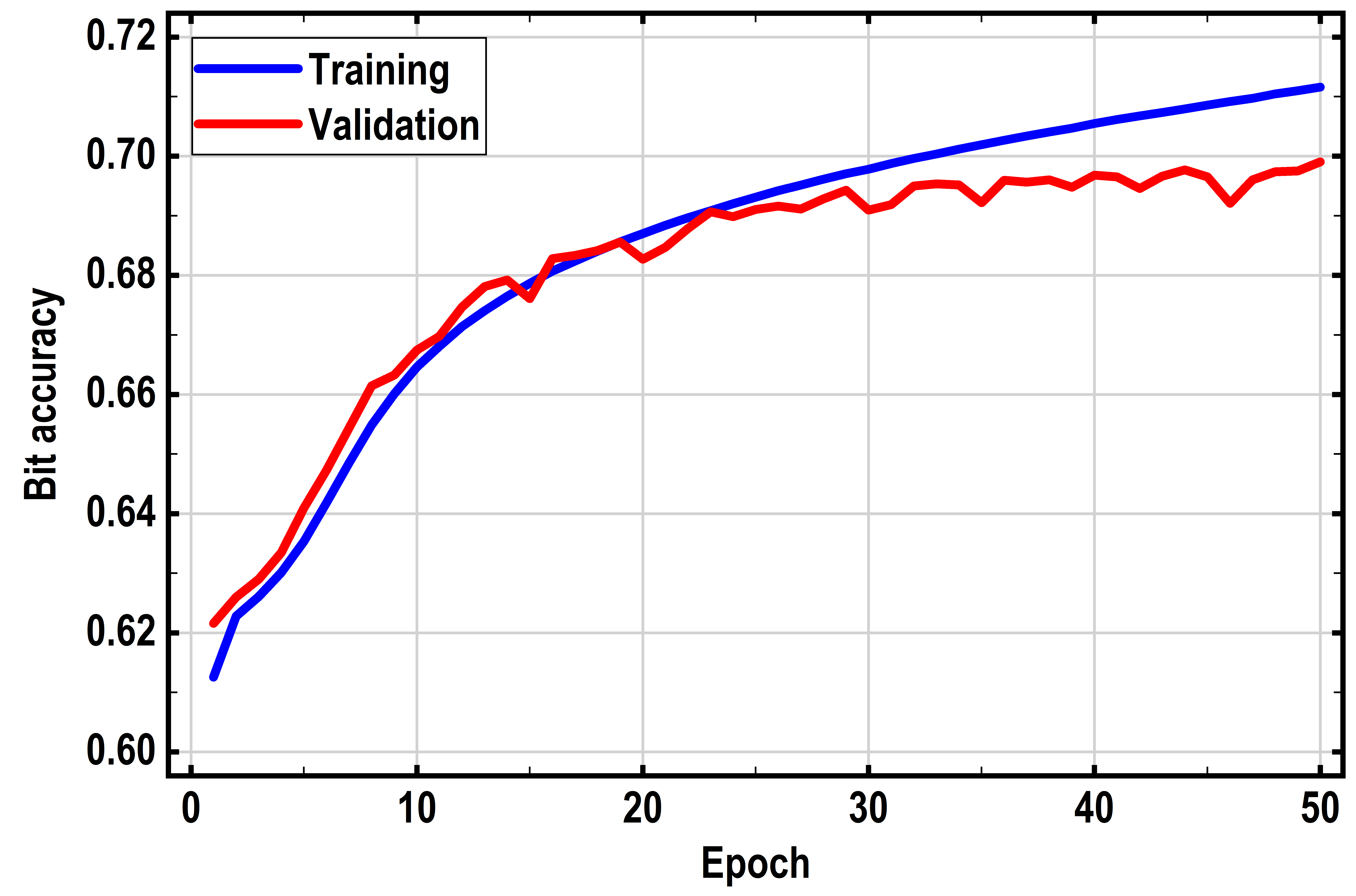}
}
\caption{Bit-level accuracy of the AR inverse model during training.}
\label{fig:bit_accuracy}
\end{figure}

Bit-level accuracy is defined as the ratio of correctly predicted pixels in the binary layout sequence and is used to evaluate how accurately the inverse model predicts each pixel position. As shown in Fig.~\ref{fig:bit_accuracy}, the bit-level accuracy progressively improves during training and converges to a validation accuracy of approximately 0.70 as the AR inverse model learns the sequential pixel distribution of the antenna layouts. This indicates that the model gradually captures the conditional relationship between the target spectrum and the sequential generation of the corresponding pixel layout.

Following the canonicalization stage in Fig.~\ref{fig:final_inverse}, the binary layout is serialized into a one-dimensional sequence using an ordering strategy designed to preserve spatial relationships between neighboring pixels in the original two-dimensional grid. Raster, snake, and Hilbert-curve orderings are compared\hypertarget{citeback:kutscher2026reordering}{}\cite{kutscher2026reordering}, and the best-performing strategy is adopted in the final model. This allows the model to learn local geometric dependencies between neighboring pixels while still modeling the global sequential dependency of the entire layout.

During training, a beginning-of-sequence (BOS) token is added at the beginning of the input sequence. The BOS token is a special start token which informs the model that pixel generation is about to begin. Since the first pixel $x_1$ has no previous pixel before it, the model uses the BOS token as the initial input when predicting $x_1$. The training input sequence is therefore constructed as
\[
[{\rm BOS}, x_1, x_2, \dots, x_{99}],
\]
and the corresponding target sequence is
\[
[x_1, x_2, x_3, \dots, x_{100}].
\]
In other words, the model learns to predict the next pixel at each sequence position. For example, the BOS token is used to predict $x_1$, the previous sequence $[{\rm BOS}, x_1]$ is used to predict $x_2$, and the previous sequence $[{\rm BOS}, x_1, x_2, \dots, x_{99}]$ is used to predict the final pixel $x_{100}$.

\begin{figure}[t]
    \centering
    \setlength{\fboxsep}{1pt}
    \setlength{\fboxrule}{0.8pt}
    \fcolorbox{white}{white}{\includegraphics[width=0.98\linewidth]{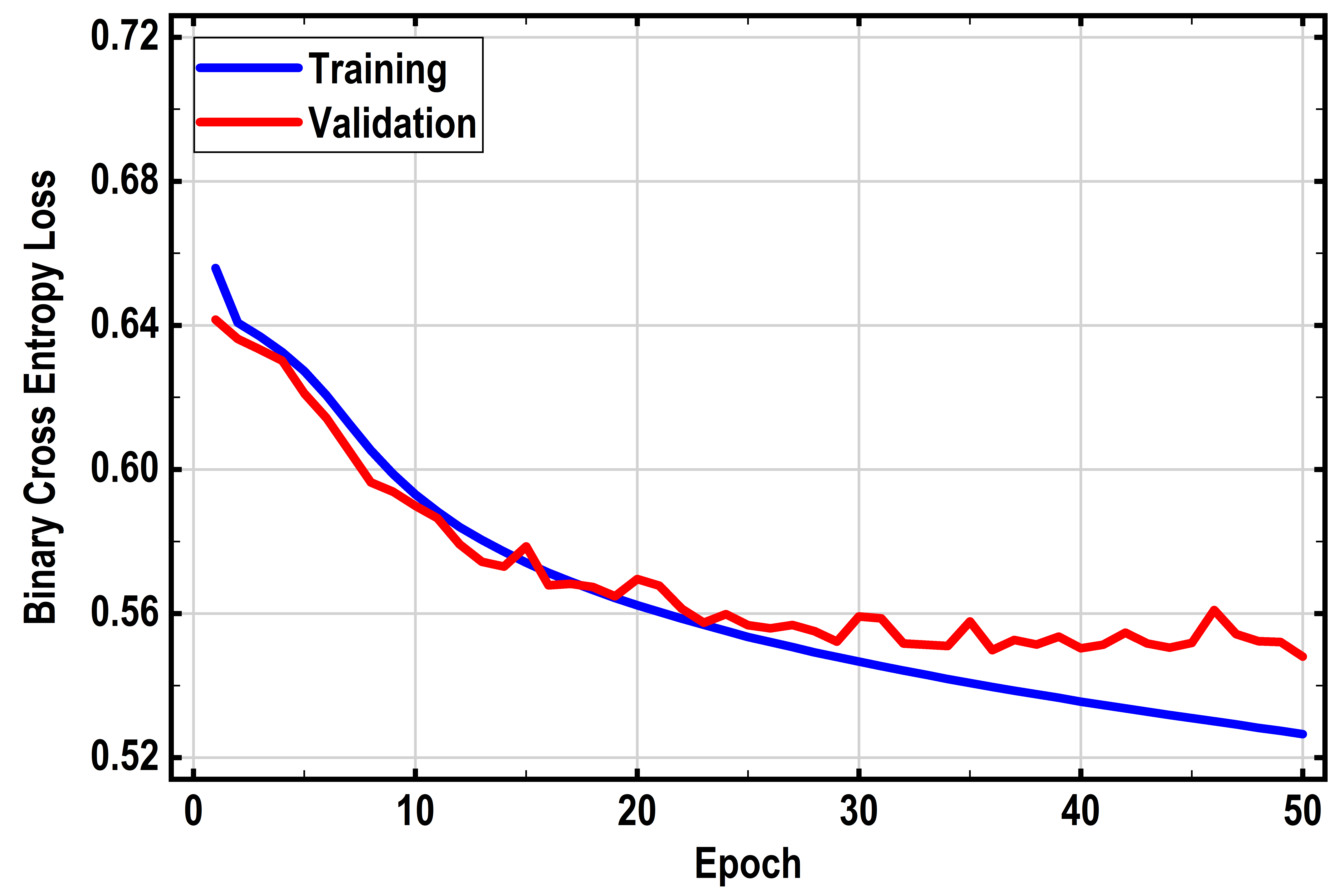}
    }
    \caption{BCE losses of the AR inverse model.}
    \label{fig:bce_loss}
\end{figure}

The serialized input sequence is represented by the sum of token embeddings, positional embeddings, and the spectral condition vector. The token embedding represents whether the previous input token corresponds to the presence or absence of a
metallic patch element, or to the BOS token. The positional embedding provides information about the location of each token in the serialized sequence. The spectral condition vector is added to every token position so that each pixel prediction is conditioned on the same target $S_{11}$ spectrum.

The resulting embeddings are processed by 15 stacked causal transformer layers with a 512-dimensional hidden representation. Each layer employs 4-head self-attention to capture dependencies among previously generated pixels from multiple representation subspaces. The self-attention output is then refined through residual connection, layer normalization, and a position-wise feed-forward network. In the trained model, the feed-forward network has an internal dimension of 768, allowing the 512-dimensional token representation at each pixel position to be expanded and nonlinearly transformed before being mapped back to the transformer hidden dimension. This stacked causal structure enables the model to generate each pixel based on the target spectrum and the previously generated pixels, without using information from future pixel positions\hypertarget{citeback:vaswani2017attention}{}\cite{vaswani2017attention}.

The model is trained by minimizing the binary cross-entropy (BCE) loss\hypertarget{citeback:nelder1972generalized}{}\cite{nelder1972generalized} between the predicted bit probabilities and the ground-truth binary pixel sequence. As shown in Fig.~\ref{fig:bce_loss}, the training loss, represented by the blue curve, steadily decreases over 50 epochs, while the validation loss, represented by the red curve, decreases from approximately 0.64 to 0.55. These trends indicate that the autoregressive model successfully learns the conditional pixel-generation process and generalizes to unseen validation samples. The training was performed on an NVIDIA RTX 4000 Ada Generation GPU and took approximately 18 hours.

For dual-band synthesis, the autoregressive model is further fine-tuned using step-function-like target responses with abrupt spectral transitions. This is motivated by the practical observation that users often know the desired operating bands, but may not know the exact spectral transition shape, notch profile, or perturbation level that should be embedded in the target response\hypertarget{citeback:li2025demand}{}\cite{li2025demand}. By representing the design objective with simplified step-function-like targets, the proposed approach allows users to specify the operating bands more intuitively, without manually constructing detailed $S_{11}$ curves. As a result, the model can be adapted to dual-band inverse design in a more practical and user-friendly manner while maintaining the sequential generation framework learned from the initially trained model.

\begin{figure}[t]
    \centering
    \setlength{\fboxsep}{1pt}
    \setlength{\fboxrule}{0.8pt}
    \fcolorbox{white}{white}{\includegraphics[width=0.98\linewidth]{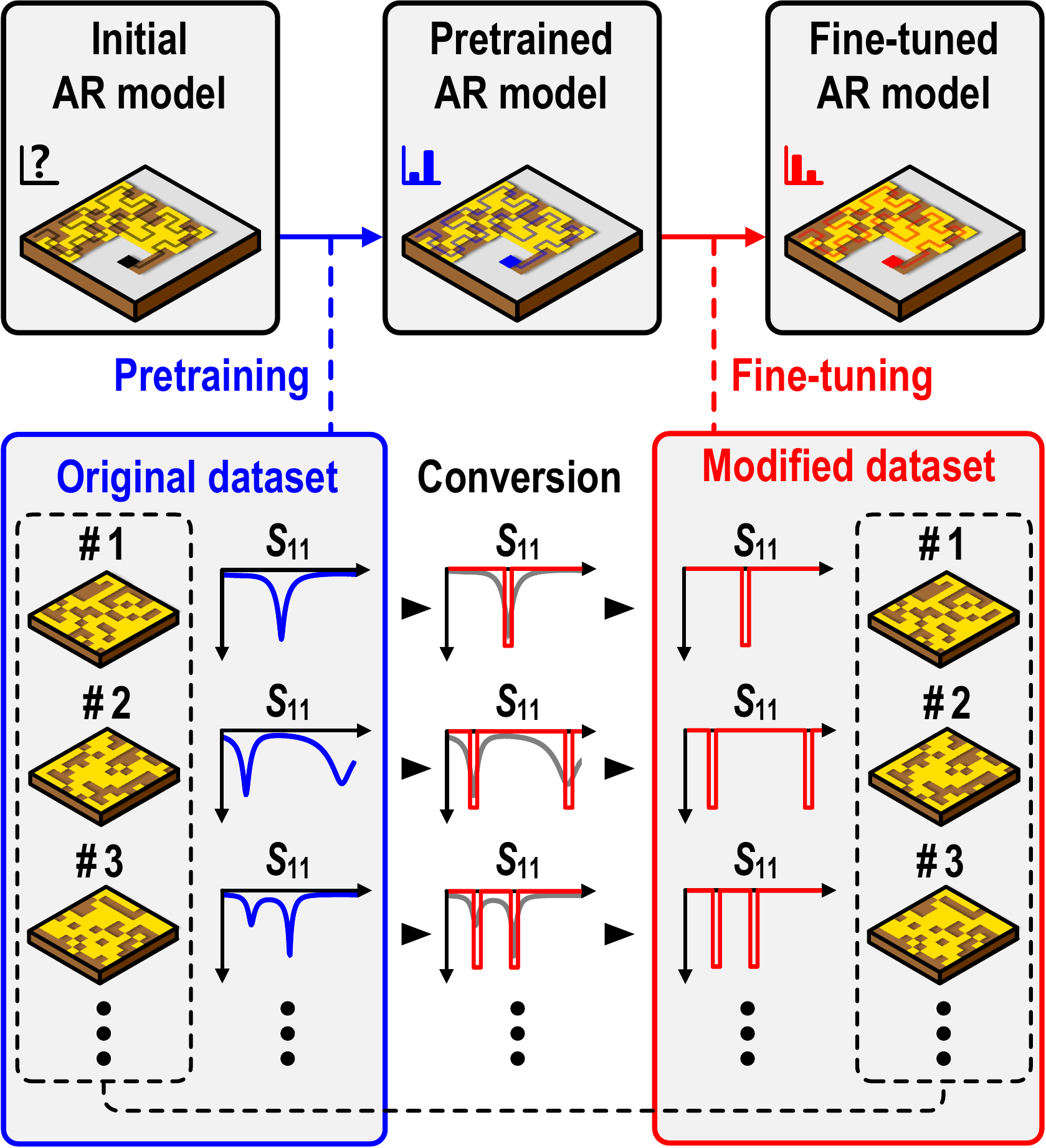}
    }
    \caption{Fine-tuning process of the AR inverse model.}
    \label{fig:finetuning_dataset}
\end{figure}

Fig.~\ref{fig:finetuning_dataset} illustrates the fine-tuning process of the proposed AR inverse model. From the original HFSS dataset, antenna layouts exhibiting single- or dual-band resonance characteristics are selected. The corresponding $S_{11}$ responses are then converted into simplified step-function targets while the associated pixel layouts remain unchanged. For each resonance band, a rectangular notch with a fixed width of 100~MHz and a magnitude of $-20$~dB is assigned, while the remaining frequency region is set to 0~dB. To improve robustness to small variations in the target-frequency specification, the step window is not always centered exactly at the original resonance minimum. Instead, its position is randomly perturbed such that the original resonance minimum lies within the central 25\% of the step window.

The pretrained AR transformer is subsequently adapted to the resulting step targets. Since the pretrained model has already learned the autoregressive dependencies among antenna pixels from continuous $S_{11}$ responses, fine-tuning is performed using different learning rates for the spectral condition encoder and the AR generation body. Specifically, a larger learning rate of $2\times10^{-5}$ is applied to the condition encoder to accommodate the domain shift from continuous spectra to simplified step representations, whereas a smaller learning rate of $5\times10^{-6}$ is applied to the pretrained Transformer and output layers to preserve the previously learned pixel-generation knowledge. The same canonicalization and snake serialization scheme used during pretraining is retained.

The model is fine-tuned using the BCE loss for up to 100 epochs with a batch size of 128. A 2-epoch learning-rate warm-up followed by cosine decay is employed, together with gradient clipping of 1.0 and early stopping with a patience of 15 epochs. The training converged through early stopping at epoch 37 and required approximately one hour. This fine-tuning strategy enables the pretrained AR model to adapt from original spectral conditioning to intuitive step specifications.

\begin{figure}[t]
    \centering
    \includegraphics[width=\linewidth]{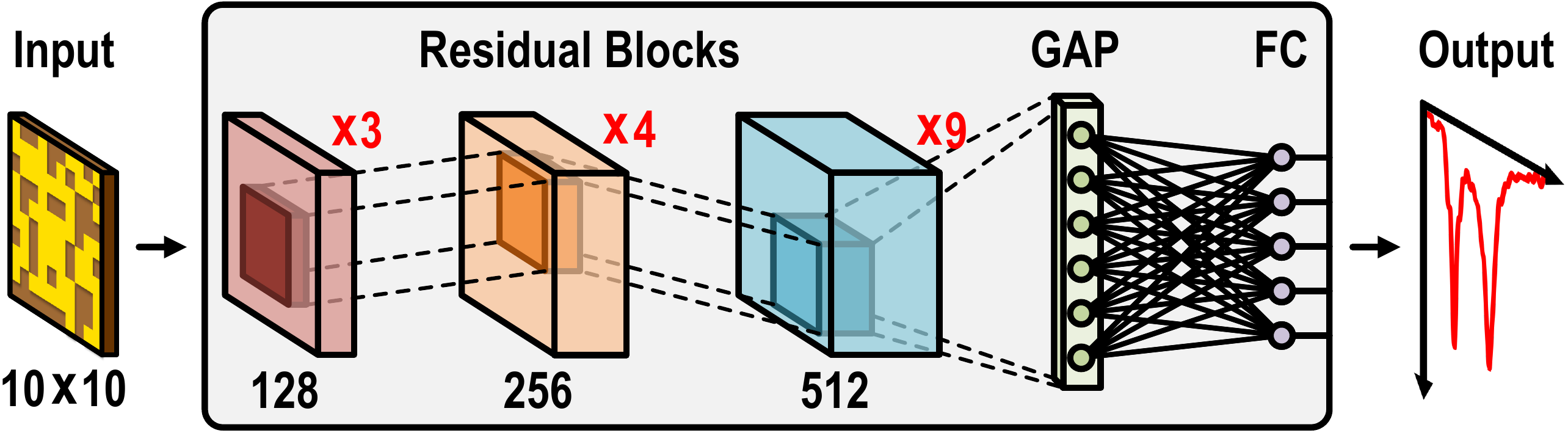}
    \caption{Forward surrogate model architecture.}
    \label{fig:forward}
\end{figure}

\subsection{Forward Surrogate Model}

In contrast to the inverse problem, the forward problem aims to predict the $S_{11}$ spectrum from a given binary pixel layout. Since this task is a deterministic mapping from geometry to response, the main requirement is to efficiently extract useful local spatial features from the layout. For this reason, a residual convolutional neural network (CNN)\hypertarget{citeback:lathuiliere2019comprehensive}{}\cite{lathuiliere2019comprehensive} is employed as the forward surrogate model.

As shown in Fig.~\ref{fig:forward}, the input binary layout is represented as a single-channel $10 \times 10$ image. The network consists of multiple residual convolutional stages, where each residual block contains two $3 \times 3$ convolution layers, batch normalization, and nonlinear activation, together with a projection shortcut when the channel dimension changes. Through these stacked residual blocks, the model learns hierarchical spatial features from the pixelated antenna geometry. After feature extraction, adaptive global average pooling (GAP) is applied, and the resulting global feature vector is mapped to the full $S_{11}$ spectrum through a fully connected (FC) layer.

The motivation for using a CNN in the forward surrogate model is fundamentally different from that of the transformer in the inverse model. In the forward problem, useful local geometric patterns that appear in one region of the pixel layout are also expected to be useful in other regions. Convolution is well suited to this role because it exploits local receptive fields and shared filters, thereby providing translation-aware feature extraction with relatively high parameter efficiency. Therefore, while the inverse model requires explicit modeling of pixel-to-pixel dependencies and conditional joint probability, the forward model mainly requires robust extraction of repeated local spatial features for deterministic spectrum regression.

\begin{figure}[!t]
    \centering
    \setlength{\fboxsep}{1pt}
    \setlength{\fboxrule}{0.8pt}
    \fcolorbox{white}{white}{\resizebox{0.98\columnwidth}{!}{\begin{tabular}{@{}c@{\hspace{1.2em}}c@{}}
                \includegraphics[height=0.115\textheight,keepaspectratio]{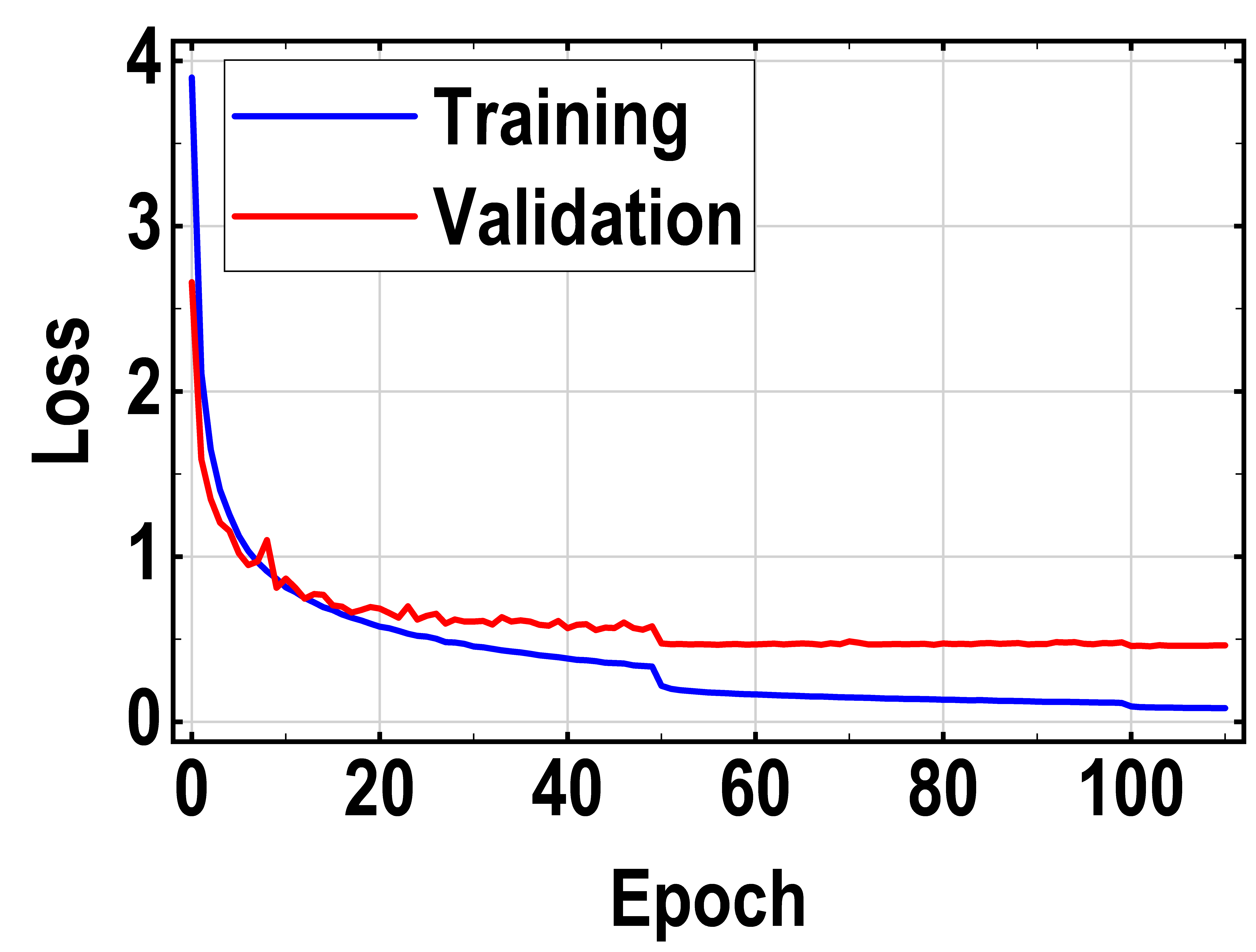}
                &
                \includegraphics[height=0.115\textheight,keepaspectratio]{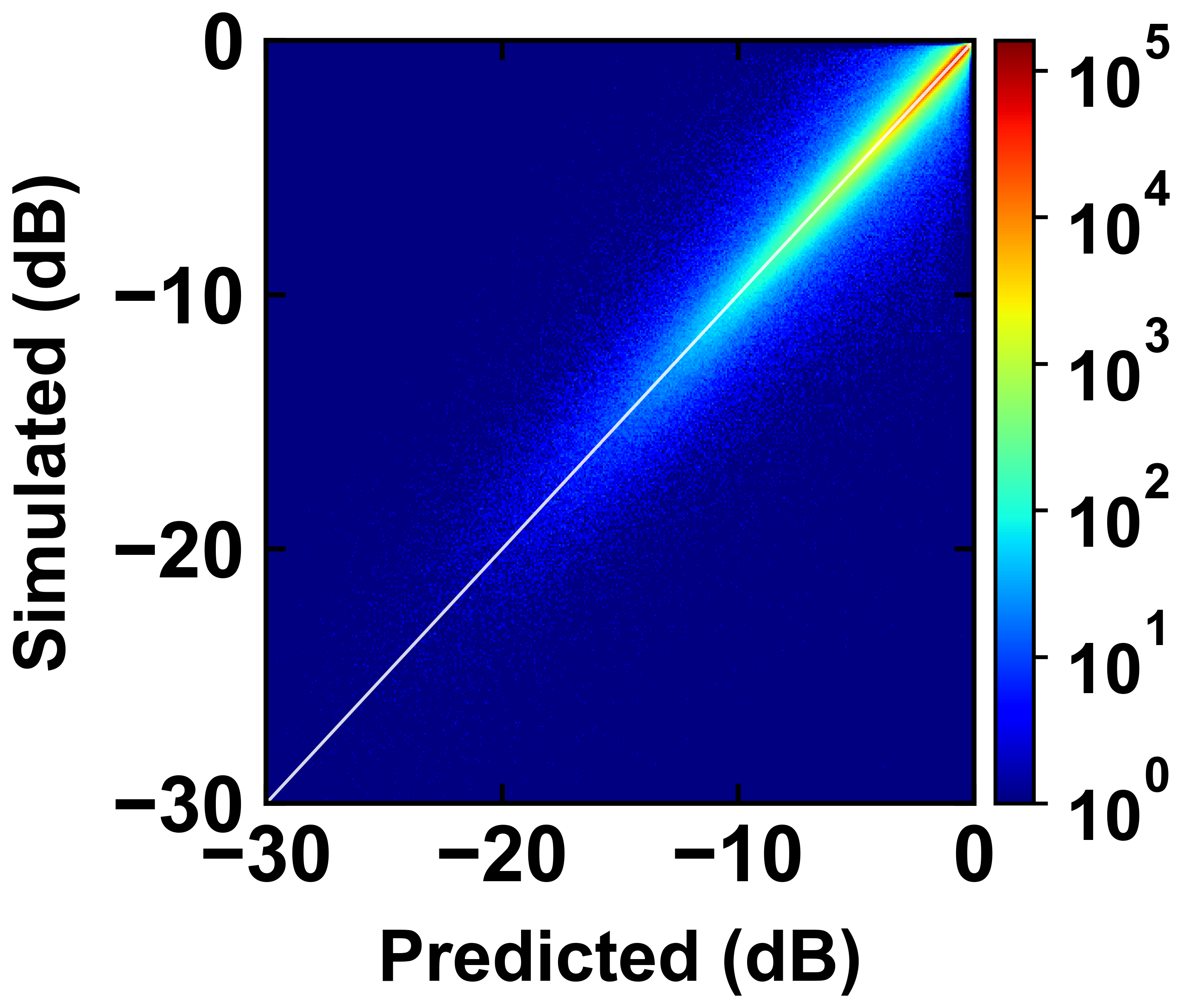}
                \\
                (a) & (b)
            \end{tabular}
        }}
    \caption{Forward surrogate model training: (a) Loss curves (b) $S_{11}$ heatmap.}
    \label{fig:forward_surrogate_training}
\end{figure}

The forward surrogate is trained in a supervised manner using the mean squared error (MSE) between the predicted and target spectra. Specifically, for a given pixelated layout, the network predicts the full $S_{11}$ response, and the MSE loss is computed over all frequency points. The model is trained for 110 epochs using the same GPU environment as the inverse network training. The AdamW optimizer is used with an initial learning rate of $1 \times 10^{-3}$ and a weight decay of $1 \times 10^{-4}$. A learning-rate decay strategy is applied to prevent the model from converging to local minima and to guide it toward the global minimum. A batch size of 128 is used during training.

The training and validation loss curves are shown in Fig.~\ref{fig:forward_surrogate_training}(a). The validation loss decreases from an initial value of 2.66 to approximately 0.46, and the training is terminated at 110 epochs as no further meaningful improvement in the validation loss is observed, indicating that the residual CNN successfully learns the mapping from the binary pixel layout to the corresponding $S_{11}$ spectrum. Although a slight degree of overfitting is observed, increasing the number of model parameters tends to improve the performance in this forward regression task. This suggests that sufficient model capacity is beneficial for capturing the complex relationship between the pixelated antenna geometry and the resulting EM response. The total training time is approximately 15 hours.

The prediction accuracy of the trained forward surrogate is further evaluated using the heatmap in Fig.~\ref{fig:forward_surrogate_training}(b), where the horizontal axis represents the predicted $S_{11}$ value and the vertical axis represents the simulated $S_{11}$ value. As shown in the heatmap, the high-density region is well aligned around this diagonal line, especially in the major response region, indicating that the model has learned the overall spectral trend sufficiently. The overall coefficient of determination reaches $R^2 = 0.92$, confirming a strong correlation between the predicted and simulated spectra. In addition, the notch-region mean absolute error (MAE) is approximately 2.13 dB, showing that the surrogate model provides a reasonably accurate approximation of the resonance-depth behavior. Once trained, the forward surrogate provides a fast approximation of the EM response of candidate pixel layouts and can therefore be used to efficiently evaluate or rank designs generated by the inverse model.

\begin{figure}[t]
    \centering
    \includegraphics[width=\linewidth]{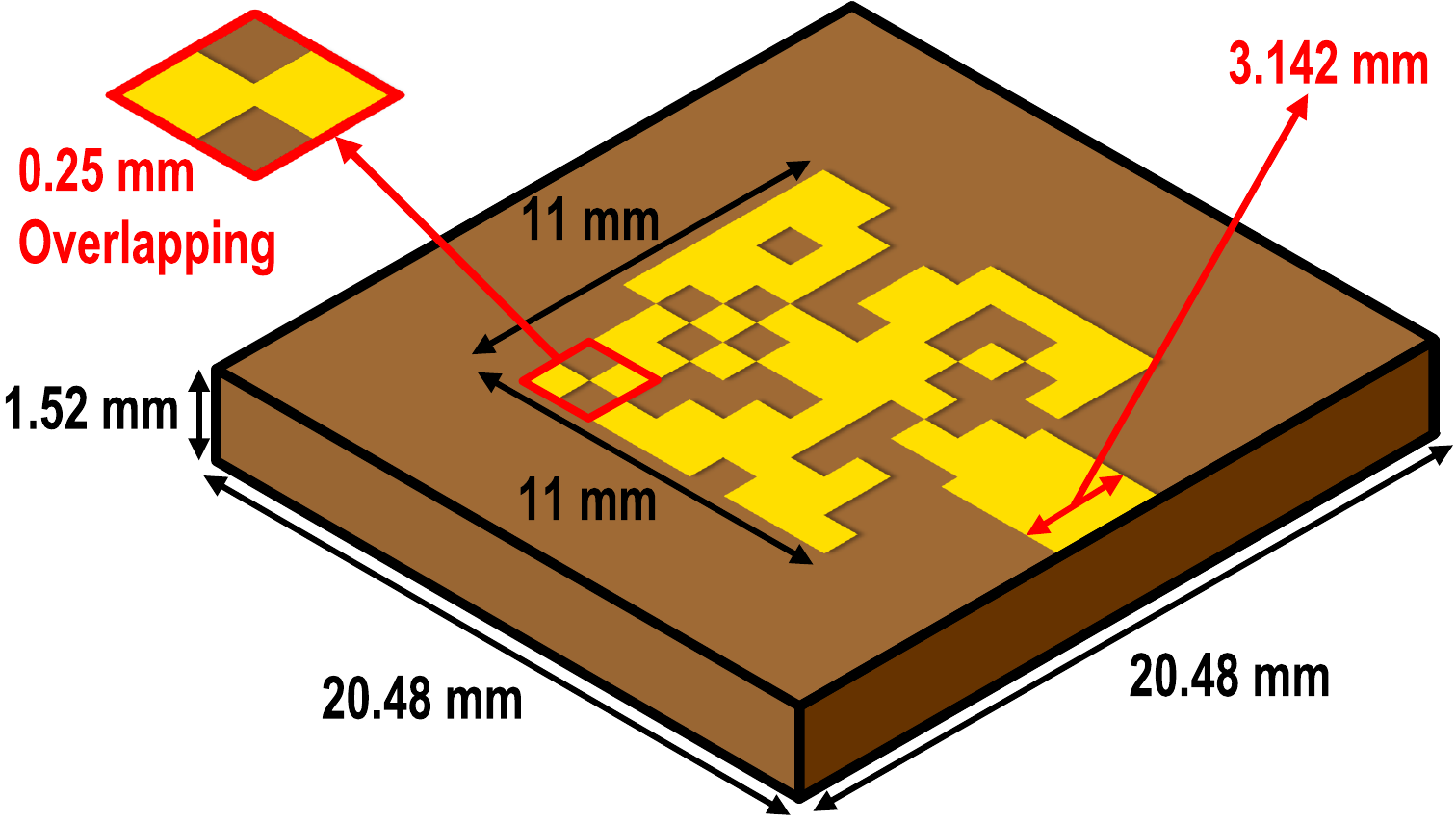}
    \caption{Pixelated patch antenna geometry.}
    \label{fig:settings}
\end{figure}

\begin{figure}[!t]
    \centering
    \setlength{\fboxsep}{1pt}
    \setlength{\fboxrule}{0.8pt}
    \fcolorbox{white}{white}{\includegraphics[width=0.98\columnwidth]{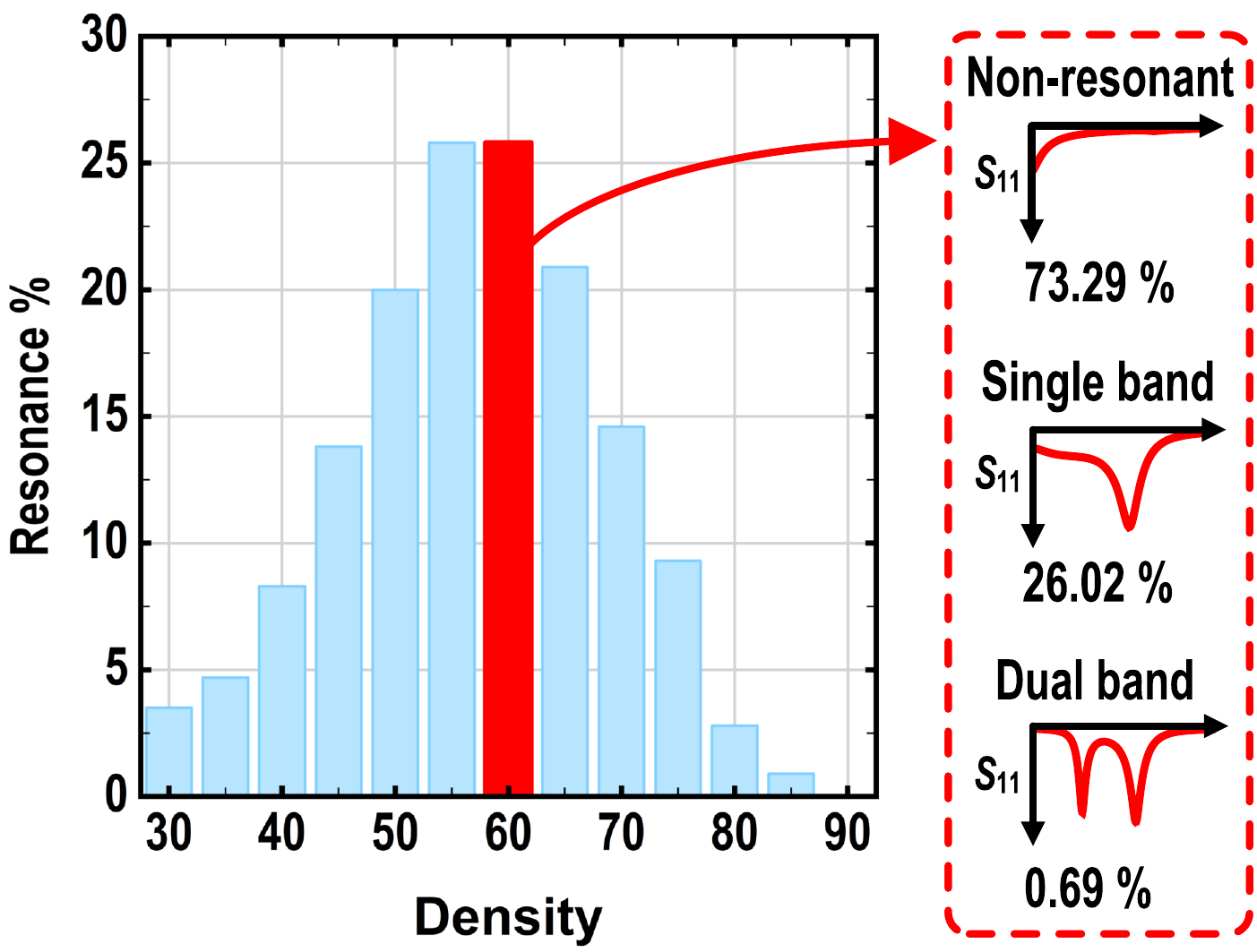}
    }
    \caption{Resonance ratio by density and spectrum distribution at 60\% density.}
    \label{fig:density_resonance}
\end{figure}

\section{Implementation Details}\label{sec:implementation}

\subsection{EM Simulation Setup and Dataset Generation}

To construct the pixel--spectrum paired dataset, full-wave EM simulations are performed using the pixelated patch antenna structure shown in Fig.~\ref{fig:settings}. The antenna is designed around a center frequency of 6 GHz using Rogers 4350B as the substrate material, with a relative dielectric constant of $\epsilon_r = 3.66$ and a thickness of 1.52 mm. The overall substrate size is set to $20.48~\mathrm{mm} \times 20.48~\mathrm{mm}$, and the pixelated patch region is defined as an $11~\mathrm{mm} \times 11~\mathrm{mm}$ square region. This patch size is determined through a preliminary patch-size scaling analysis, in which 1000 EM simulations are conducted for each scaling condition to form a resonance near 6 GHz. The feed width of 3.142 mm is employed to achieve 50-$\Omega$ characteristic impedance. Note that an overlap of 0.25 mm is introduced between neighboring pixels to ensure electrical connectivity~\cite{chen2025inverse},\hypertarget{citeback:wakrim2017multiband}{}\cite{wakrim2017multiband},\hypertarget{citeback:wakrim2021novel}{}\cite{wakrim2021novel}.

To select an appropriate pixel density for dataset generation, the resonance occurrence rate is analyzed under different density conditions, as shown in Fig.~\ref{fig:density_resonance}. Similar to the patch-size scaling analysis, 1000 EM simulations are conducted for each density condition. Among the tested density values, the 60\% density condition exhibits one of the highest resonance ratios. Therefore, this density is selected to increase the number of resonant antenna patterns in the dataset, which is important for enabling the model to effectively learn meaningful notch responses\hypertarget{citeback:khaireh2023newcomer}{}\cite{khaireh2023newcomer}. Based on this condition, each pixelated antenna sample is generated by randomly selecting 60 active pixels among the 100 pixel positions.

For each generated layout, a Python script automatically
controls the HFSS EM simulation process by repeatedly
executing the required simulation macro. The simulated $S_{11}$ is extracted over the frequency range from 5 to 7 GHz with
201 frequency points. The resonance statistics of the generated
60\%-density dataset are also summarized in Fig.~\ref{fig:density_resonance}. As can
be observed from the figure, 73.29\% of the generated samples
are classified as non-resonant responses, 26.02\% as single-band responses, and 0.69\% as dual-band responses.

\subsection{Comparison of Searching Algorithms}

To determine an effective inference strategy for the proposed AR transformer, three searching algorithms are compared: random sampling\hypertarget{citeback:holtzman2019curious}{}\cite{holtzman2019curious}, beam search\hypertarget{citeback:freitag2017beam}{}\cite{freitag2017beam}, and multi-armed bandit (MAB)\hypertarget{citeback:auer2002finite}{}\cite{auer2002finite} based search. Random sampling generates antenna layouts probabilistically from the pixel-wise output distribution of the AR transformer, which provides high candidate diversity but does not directly guide the generation toward high-performance antenna responses. Beam search retains multiple high-probability generation paths during autoregressive decoding, allowing likely layouts to be generated efficiently. However, because it repeatedly favors high-likelihood paths, the generated candidates can become less diverse\hypertarget{citeback:ippolito2019comparison}{}\cite{ippolito2019comparison},\hypertarget{citeback:shi2024thorough}{}\cite{shi2024thorough}.

\begin{figure}[!t]
    \centering
    \setlength{\fboxsep}{1pt}
    \setlength{\fboxrule}{0.8pt}
    \fcolorbox{white}{white}{\includegraphics[width=0.98\columnwidth]{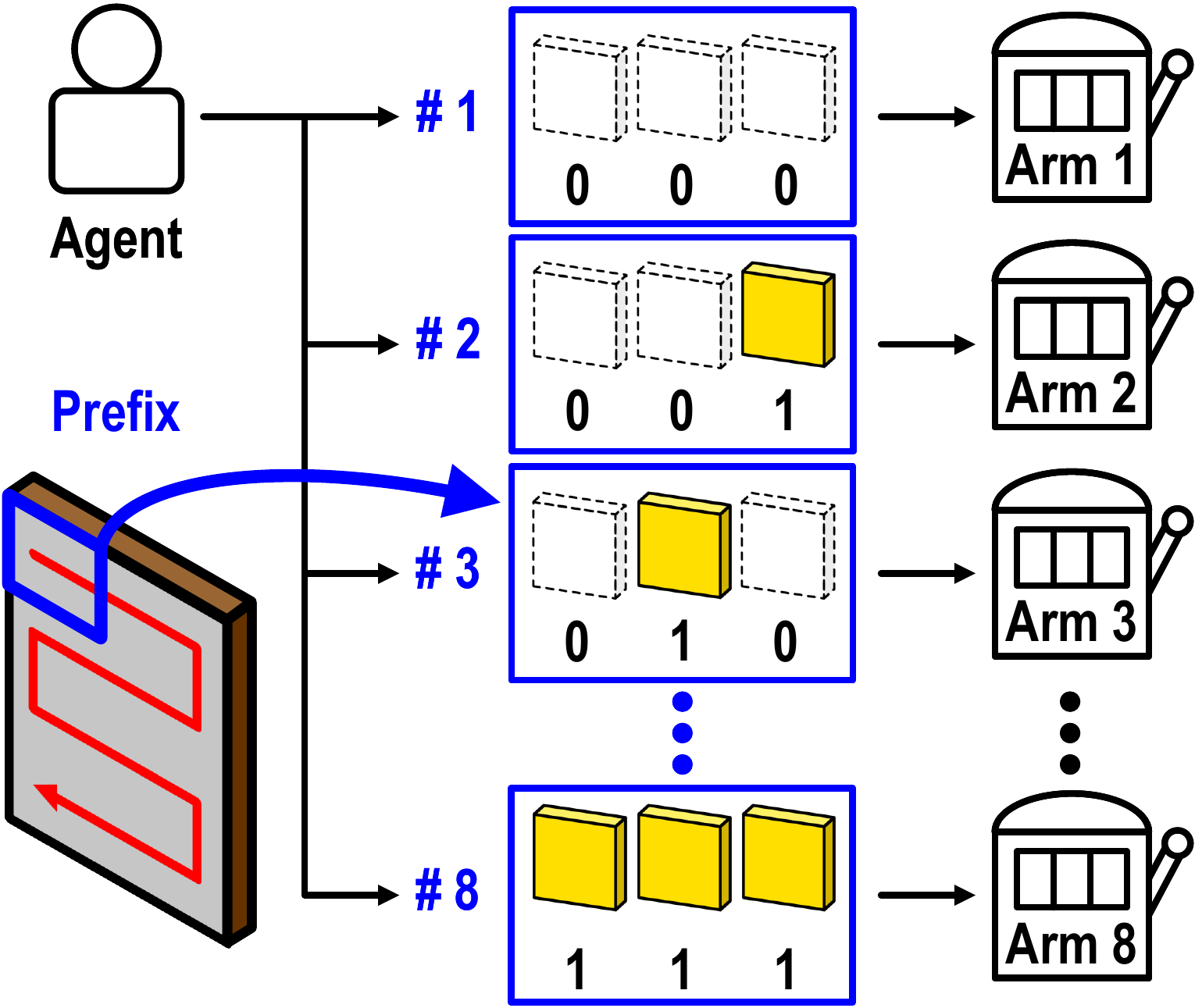}
    }
    \caption{Conceptual illustration of the multi-armed bandit (MAB) problem.}
    \label{fig:multi_armed_bandit}
\end{figure}

The MAB-based search is adopted to guide candidate generation more adaptively. As illustrated in Fig.~\ref{fig:multi_armed_bandit}, the general MAB problem consists of an agent and multiple selectable arms. At each trial, the agent selects one arm, observes the corresponding reward, and updates its future selection strategy based on the accumulated rewards. This framework is useful for balancing exploration of less-tested options and exploitation of options that have already shown favorable results\hypertarget{citeback:audibert2009exploration}{}\cite{audibert2009exploration},\hypertarget{citeback:carpentier2011upper}{}\cite{carpentier2011upper}. In this work, the MAB concept is applied to the prefix selection of the AR transformer. Several initial binary prefix patterns are treated as selectable arms, and the remaining pixels are generated by stochastic autoregressive sampling. Therefore, the MAB-based search adaptively allocates more trials to prefix regions that are likely to produce high-quality antenna candidates. The upper confidence bound criterion is used to balance exploration and exploitation.

Each generated candidate is evaluated using the forward surrogate model, and the candidate MSE is computed over the full $S_{11}$ spectrum. A candidate is regarded as valid when its predicted $S_{11}$ satisfies the target notch condition at $-12$~dB, stricter than the conventional $-10$~dB, to ensure a practical bandwidth margin. To quantitatively compare the three searching algorithms, the same 334 unseen dual-band test cases are used for comparison. Since multiple algorithms can find valid candidates for the same target, a simple binary criterion based only on whether a valid candidate is found is insufficient. Therefore, each algorithm is evaluated using the best candidate found within the same 600-second search budget. For each target case, the lowest full-band MSE among the valid candidates is used as the representative performance, and the algorithms were compared using this metric. A lower MSE indicates that the algorithm finds a candidate whose predicted response more closely matches the target.

\begin{figure}[!t]
    \centering
    \setlength{\fboxsep}{1pt}
    \setlength{\fboxrule}{0.8pt}
    \fcolorbox{white}{white}{\resizebox{0.97\columnwidth}{!}{\begin{tabular}{@{}c@{\hspace{0.35em}}c@{\hspace{0.35em}}c@{}}
                \includegraphics[height=0.103\textheight,keepaspectratio]{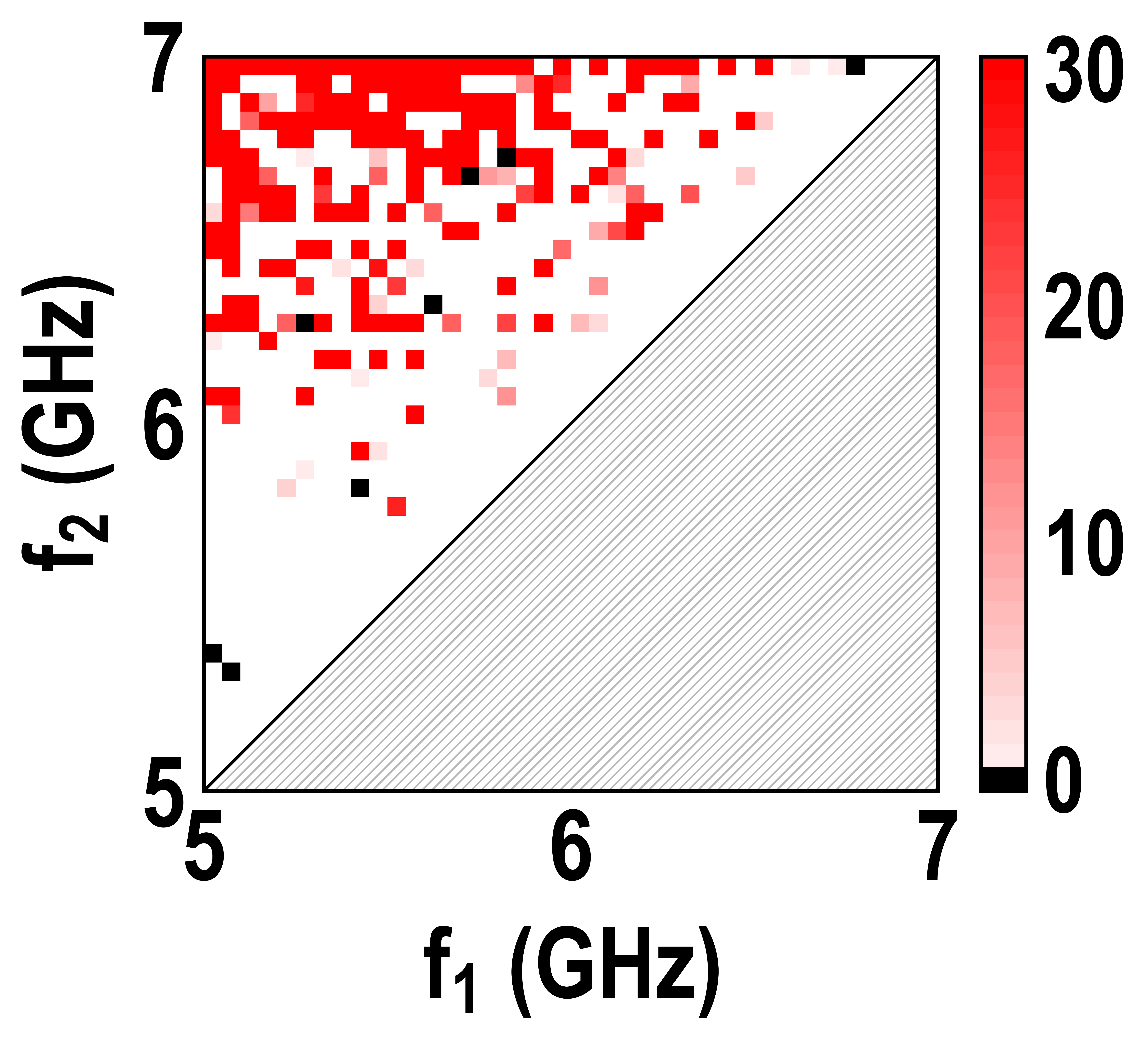}
                &
                \includegraphics[height=0.103\textheight,keepaspectratio]{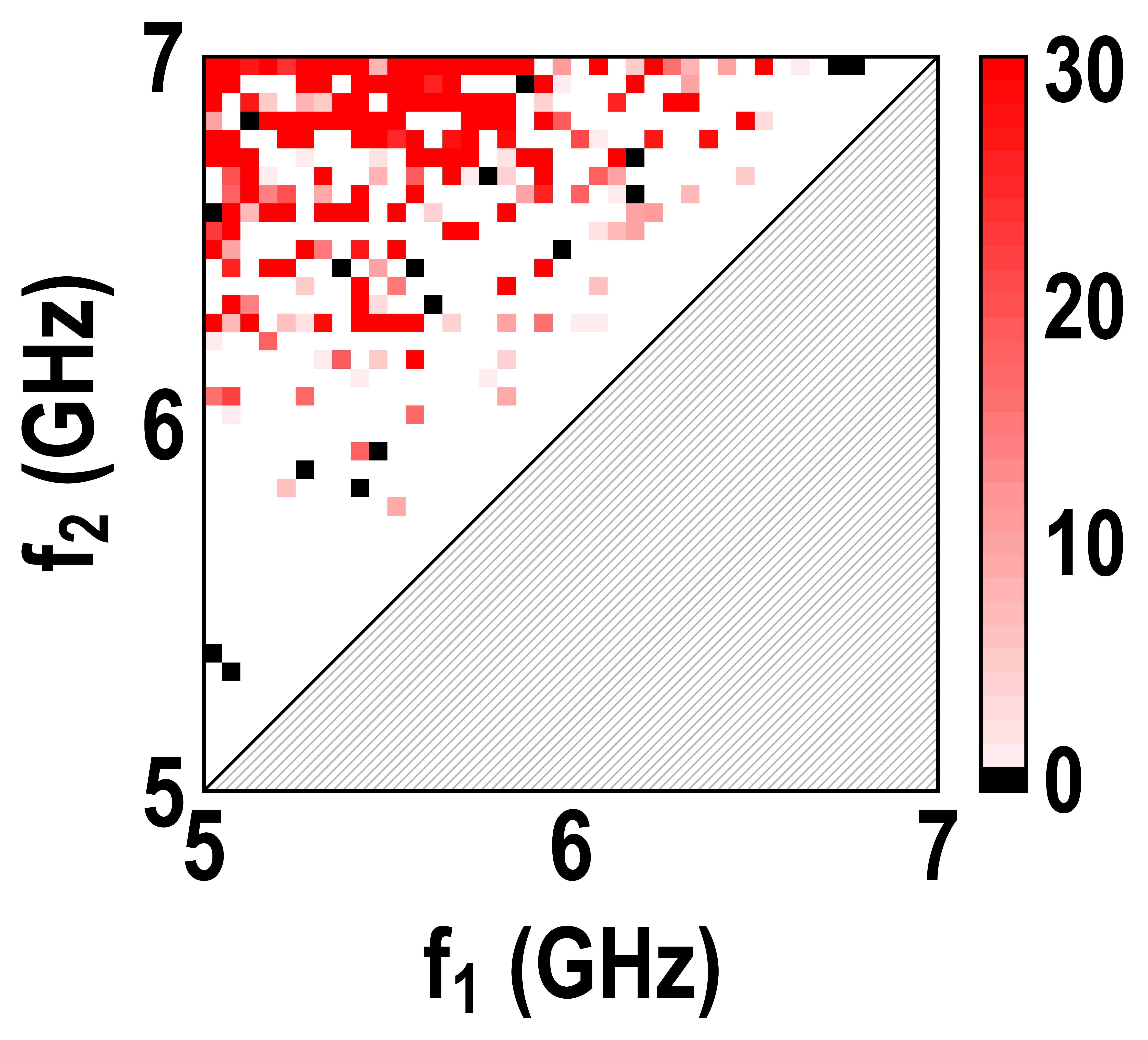}
                &
                \includegraphics[height=0.103\textheight,keepaspectratio]{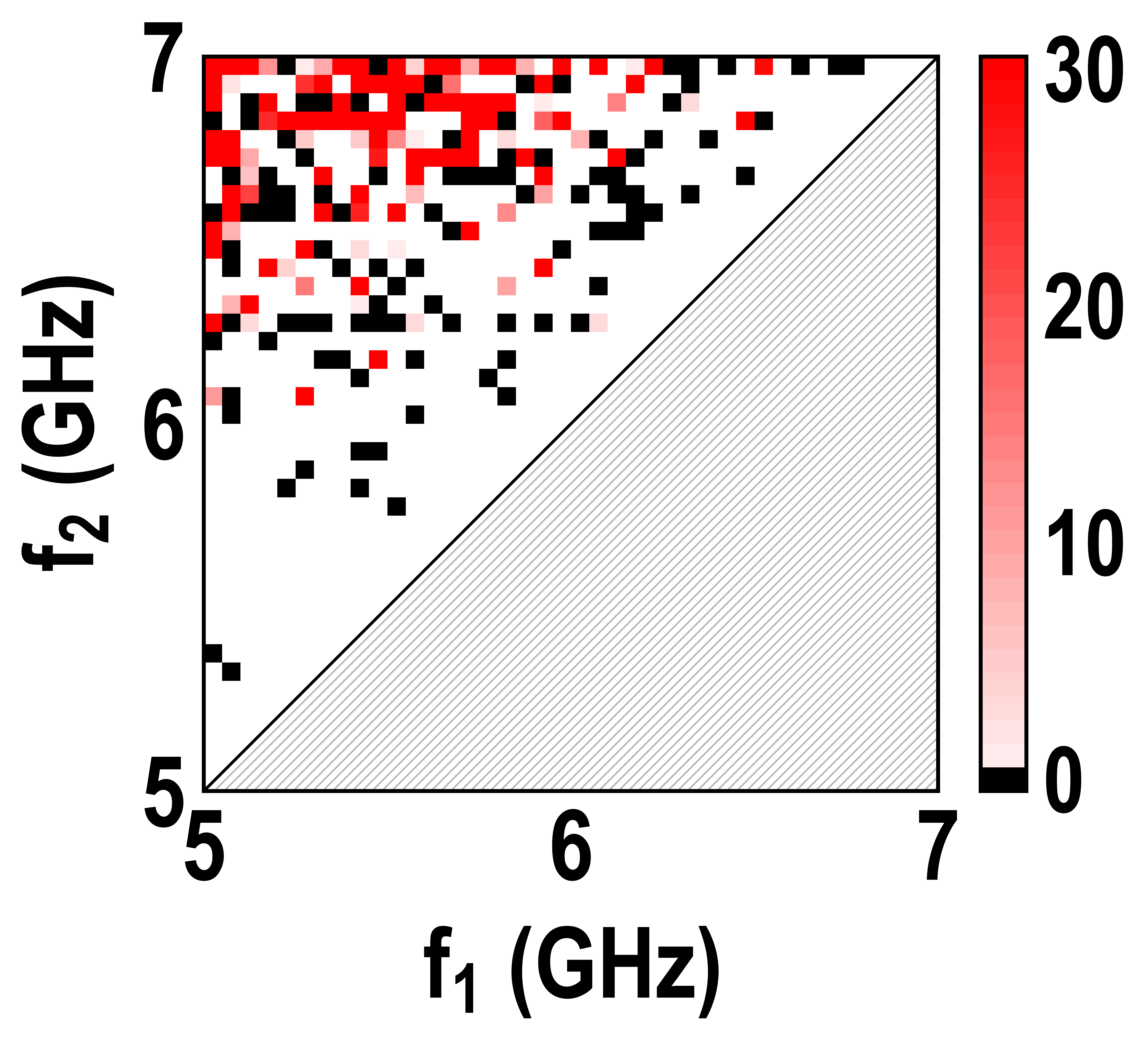}
                \\
                (a) & (b) & (c)
            \end{tabular}}}
    \caption{Heatmaps of unique valid-candidate counts for searching algorithms: (a) multi-armed bandit, (b) random sampling, and (c) beam search.}
    \label{fig:search_strategy_heatmaps}
\end{figure}

Fig.~\ref{fig:search_strategy_heatmaps} compares the number of unique valid candidates generated by the three searching algorithms for the same dual-band test cases. Here, $f_1$ and $f_2$ denote the lower and higher target resonance frequencies, respectively. Since each dual-band target is defined such that $f_1<f_2$, the lower triangular region corresponding to $f_1\geq f_2$ does not represent valid dual-band combinations and is therefore shown as a hatched region. In the valid upper triangular region, a higher candidate count indicates that a larger number of distinct valid antenna layouts are obtained for the corresponding target, whereas black points indicate cases in which no unique valid candidate is found. The MAB-based search exhibits broader coverage and fewer unsuccessful cases than random sampling and beam search.

\begin{table}[!t]
    \centering
    \caption{Performance comparison of searching algorithms}
    \label{tab:algo_ranking}
    \footnotesize
    \begin{tblr}{
        width = 0.98\columnwidth,
        colspec = {
            |X[1.15,c,m]
            |X[1.05,c,m]
            |X[1.45,c,m]|
        },
        hlines,
        rows = {halign=c, valign=m},
        row{1} = {font=\bfseries, ht=6.2ex},
        row{2} = {ht=3.8ex},
        row{3} = {ht=3.8ex},
        row{4} = {ht=3.8ex},
        rowsep = 0pt,
        colsep = 2pt
    }
        Algorithm
        & {Number of\\top-ranked cases}
        & {Average full-band MSE\\
           of best valid candidate} \\

        Multi-Armed Bandit
        & 234 / 334
        & 0.92 \\

        Random Sampling
        & 78 / 334
        & 1.19 \\

        Beam Search
        & 22 / 334
        & 1.67 \\
    \end{tblr}
\end{table}

Table~\ref{tab:algo_ranking} summarizes the comparison results. Here, ``Number of top-ranked cases'' denotes the number of target cases in which each algorithm achieved the lowest full-band MSE. The MAB-based search achieves the largest number of best cases, with 234 out of 334 cases. It also achieves the lowest average full-band MSE of the best valid candidate at 0.92, indicating that the MAB-based search can identify high-quality candidates more effectively than other searching algorithms.

\subsection{Comparison of BPSO and the Proposed AR Transformer}

\begin{figure*}[!t]
\centering
    \setlength{\fboxsep}{1pt}
    \setlength{\fboxrule}{0.8pt}

    \subfigure[]{\fcolorbox{white}{white}{\includegraphics[width=0.98\textwidth]{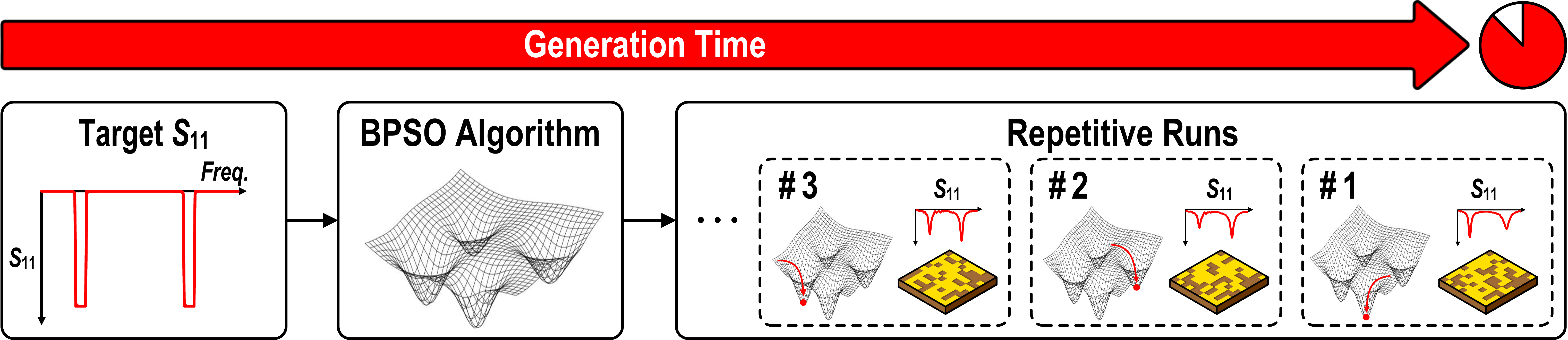}
        }}
    
    \vspace{0.3em}
    
    \subfigure[]{\fcolorbox{white}{white}{\includegraphics[width=0.98\textwidth]{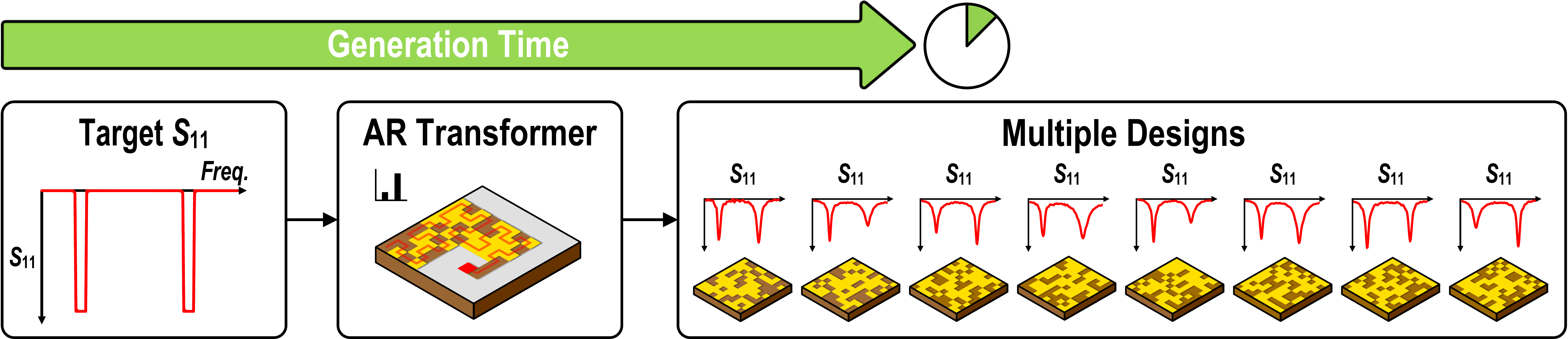}
        }}
    
\caption{Inverse design workflows of (a) BPSO-based iterative optimization and (b) AR transformer-based direct generation.}
\label{fig:generation_comparison}
\end{figure*}

To quantitatively evaluate the candidate generation capability of the proposed inverse design model, the generation performance of BPSO and the proposed AR transformer is compared. BPSO is a widely used population-based optimization algorithm for discrete design problems, but its search performance is strongly affected by the initial condition and parameter tuning\cite{kim2025inverse},\hypertarget{citeback:zheng2003convergence}{}\cite{zheng2003convergence},\hypertarget{citeback:liu2015analysis}{}\cite{liu2015analysis}. In particular, multiple independent optimization runs are often required to obtain several valid antenna geometries for a given target response, which can increase the overall generation time. Therefore, the comparison is conducted to numerically demonstrate the difference in generation efficiency.

Fig.~\ref{fig:generation_comparison} illustrates the difference between the two inverse design workflows. In the BPSO-based approach, candidate antenna geometries are obtained through iterative particle updates, and multiple optimization runs are generally required to obtain diverse valid solutions for the same target spectrum. In contrast, the proposed AR transformer directly synthesizes pixelated antenna layouts conditioned on the target $S_{11}$ response through sequential pixel generation, enabling multiple distinct candidates to be generated without repetitive optimization.

For a comprehensive quantitative comparison, the pretrained AR model is compared with conventional BPSO and its variants on 334 unseen dual-band targets from the held-out test set using the same budget of 20,000 evaluated candidates per target. To account for the parameter sensitivity of conventional BPSO, 36 hyperparameter settings are evaluated by varying the inertia weight $w$, cognitive acceleration coefficient $c_1$, and social acceleration coefficient $c_2$ within the recommended ranges based on widely adopted PSO tuning guidelines\hypertarget{citeback:innocente2010coefficients}{}\cite{innocente2010coefficients},\hypertarget{citeback:shami2022particle}{}\cite{shami2022particle},\hypertarget{citeback:rezaee2013parameter}{}\cite{rezaee2013parameter}. Each setting is evaluated over all test cases, and the settings achieving the highest and lowest coverage are reported as BPSO (best) and BPSO (worst), respectively. V-shaped BPSO (VBPSO)\cite{ullah2023multiservice} and Modified V-shaped BPSO (MVBPSO)\hypertarget{citeback:kumar2026automated}{}\cite{kumar2026automated} are additionally evaluated.

\begin{figure}[!t]
    \centering
    \setlength{\fboxsep}{1pt}
    \setlength{\fboxrule}{0.8pt}
    \fcolorbox{white}{white}{\resizebox{0.97\columnwidth}{!}{\begin{tabular}{@{}c@{\hspace{0.35em}}c@{\hspace{0.35em}}c@{}}
                \includegraphics[height=0.103\textheight,keepaspectratio]{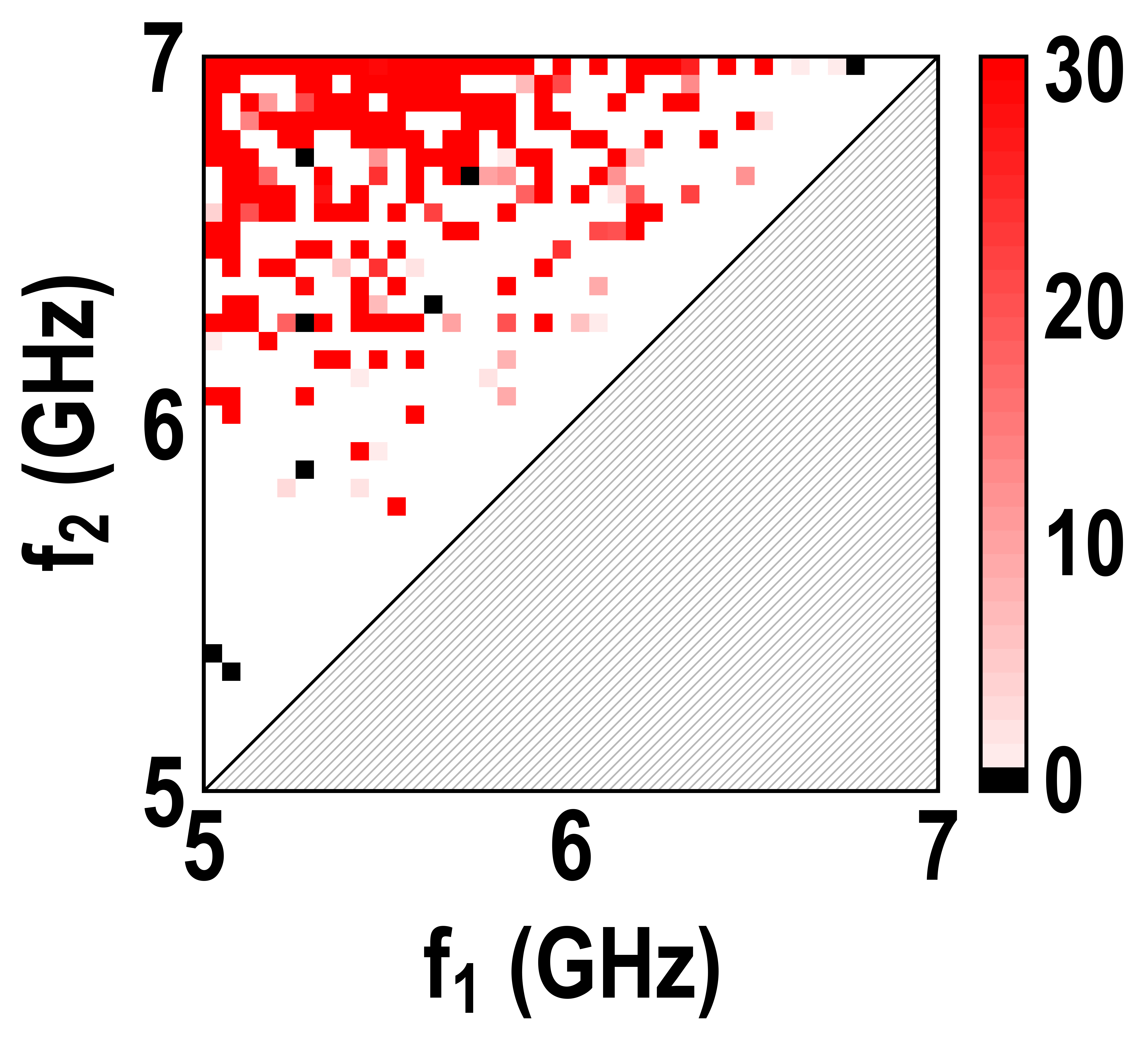}
                &
                \includegraphics[height=0.103\textheight,keepaspectratio]{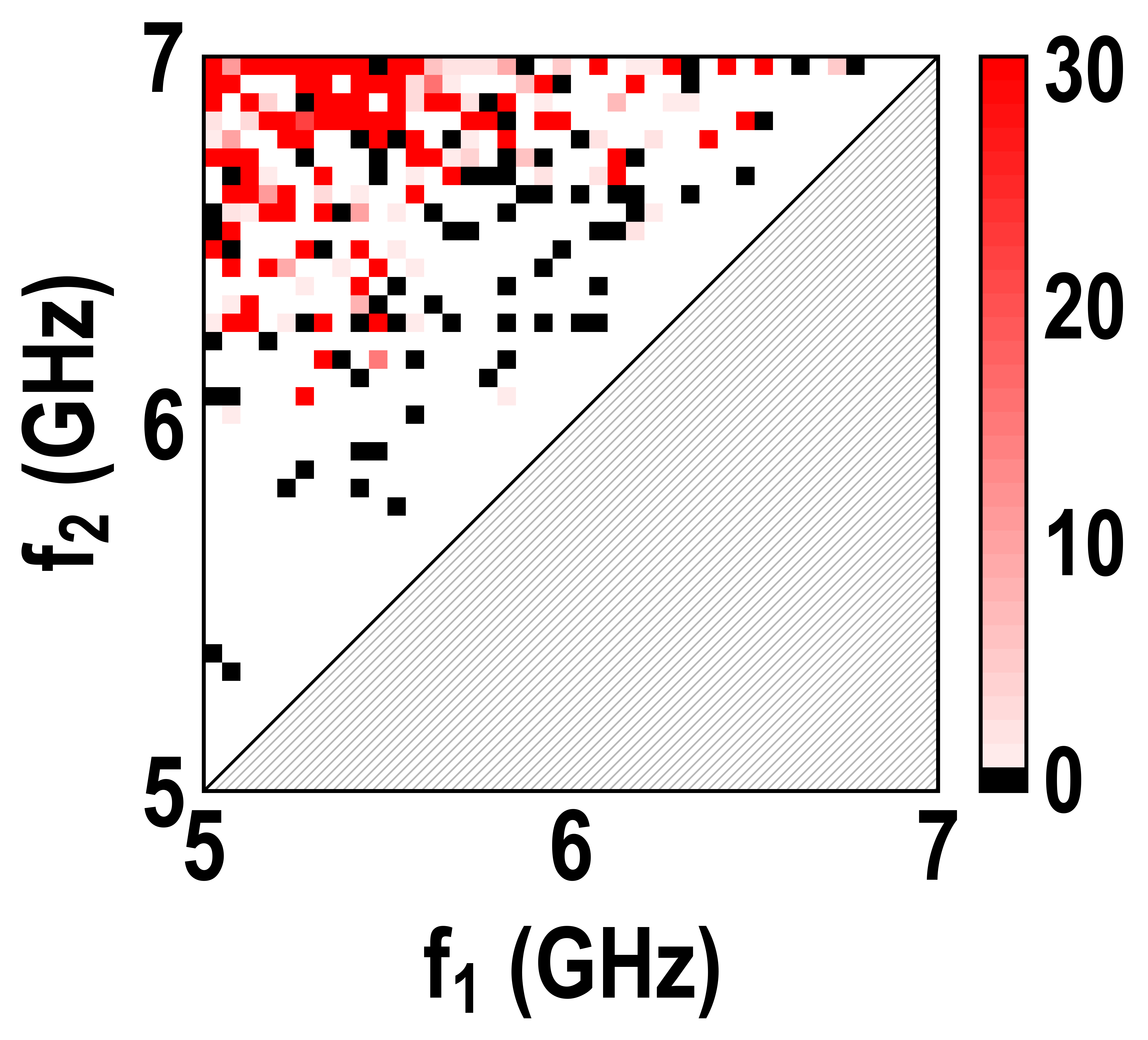}
                &
                \includegraphics[height=0.103\textheight,keepaspectratio]{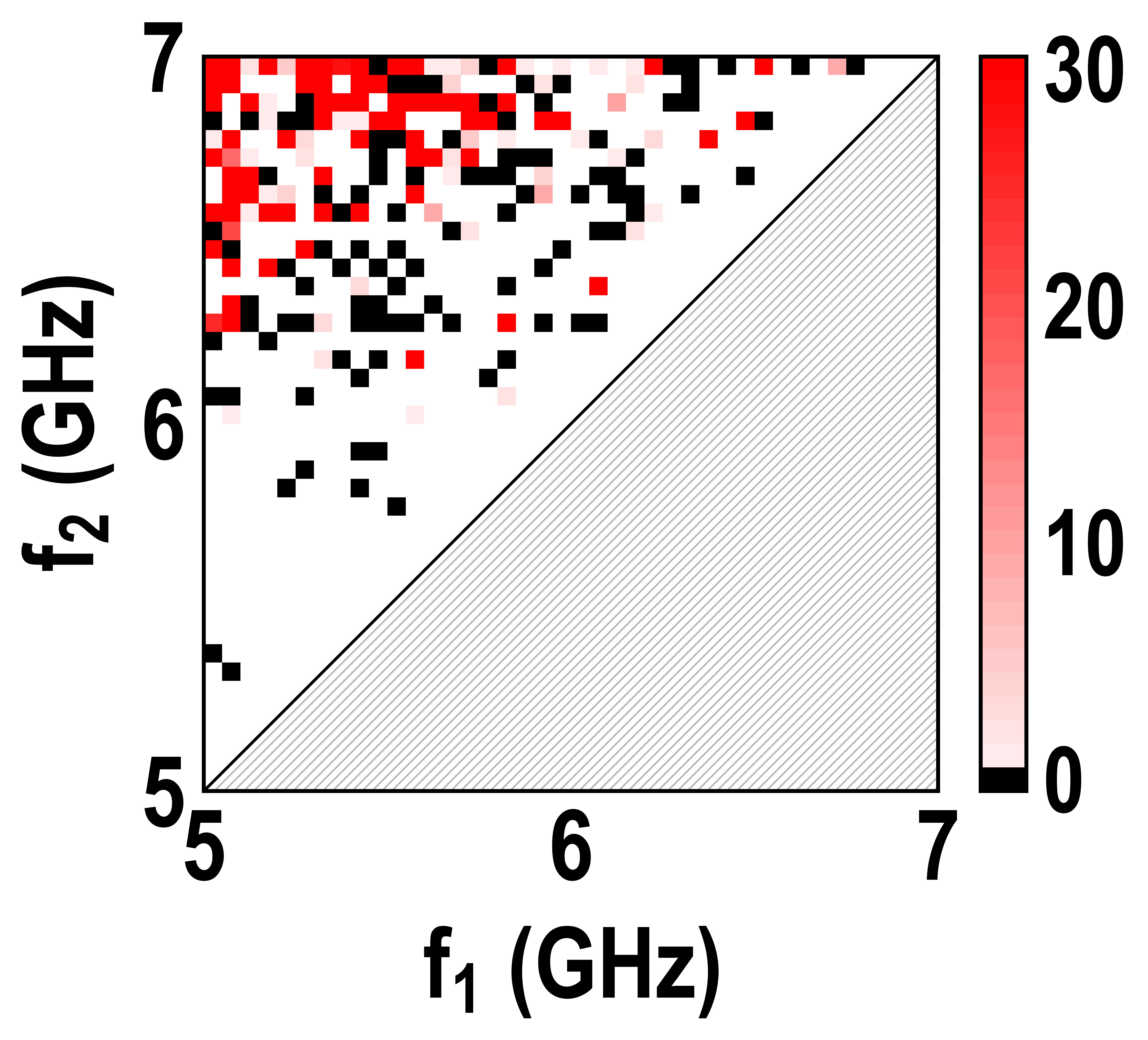}
                \\
                (a) & (b) & (c)
            \end{tabular}}}
    \caption{Heatmaps of unique valid-candidate counts over 334 dual-band test cases: (a) pretrained AR model, (b) BPSO best, and (c) BPSO worst.}
    \label{fig:pretrained_test_heatmaps}
\end{figure}

\begin{table}[!t]
    \centering
    \caption{Comparison of the pretrained AR model and BPSO variants}
    \label{tab:pretrained_bpso_coverage}
    \scriptsize
    \color{black}
    \begin{tblr}{
        width = 0.98\columnwidth,
        colspec = {
            |X[1.10,c,m]
            |X[0.90,c,m]
            |X[0.78,c,m]
            |X[1.42,c,m]
            |X[1.25,c,m]|
        },
        hlines,
        rows = {halign=c, valign=m},
        row{1} = {font=\bfseries, ht=6.4ex},
        row{2} = {ht=4.0ex},
        row{3} = {ht=4.0ex},
        row{4} = {ht=4.0ex},
        row{5} = {ht=4.0ex},
        row{6} = {ht=4.0ex},
        rowsep = 0pt,
        colsep = 2pt
    }
        Method
        & {Successful\\cases}
        & {Coverage\\(\%)}
        & {Redundant valid\\candidates (\%)}
        & {Average best\\full-band MSE} \\

        Pretrained AR
        & 322/334
        & 96.4
        & 0.0
        & 0.89 \\

        BPSO (best)
        & 202/334
        & 60.5
        & 20.4
        & 7.67 \\

        BPSO (worst)
        & 171/334
        & 51.2
        & 34.2
        & 9.05 \\

        VBPSO
        & 184/334
        & 55.1
        & 13.8
        & 4.18 \\

        MVBPSO
        & 192/334
        & 57.5
        & 71.3
        & 3.50 \\
    \end{tblr}
\end{table}

As summarized in Fig.~\ref{fig:pretrained_test_heatmaps} and Table~\ref{tab:pretrained_bpso_coverage}, the pretrained AR model provides substantially broader target-space coverage than conventional BPSO, VBPSO, and MVBPSO, while producing non-redundant valid solutions. It also achieves the lowest average best full-band spectrum MSE among the evaluated methods, demonstrating more accurate and diverse multi-solution generation across the dual-band target space.

\subsection{Fine-Tuned Model Evaluation}

The fine-tuned model is evaluated using 153 dual-band step-function targets with band centers from 5.1 to 6.9~GHz at 0.1-GHz intervals. Adjacent centers are excluded so that two 100-MHz notches remain separated; each notch is fixed at $-20$~dB with 0~dB elsewhere. As in the pretrained-model evaluation, the 36 BPSO hyperparameter settings are evaluated over all 153 targets, and the settings with the highest and lowest coverage are reported as BPSO (best) and BPSO (worst), respectively. VBPSO and MVBPSO are additionally evaluated using the same target set and candidate budget.

\begin{figure}[!t]
    \centering
    \setlength{\fboxsep}{1pt}
    \setlength{\fboxrule}{0.8pt}
    \fcolorbox{white}{white}{\resizebox{0.97\columnwidth}{!}{\begin{tabular}{@{}c@{\hspace{0.35em}}c@{\hspace{0.35em}}c@{}}
                \includegraphics[height=0.103\textheight,keepaspectratio]{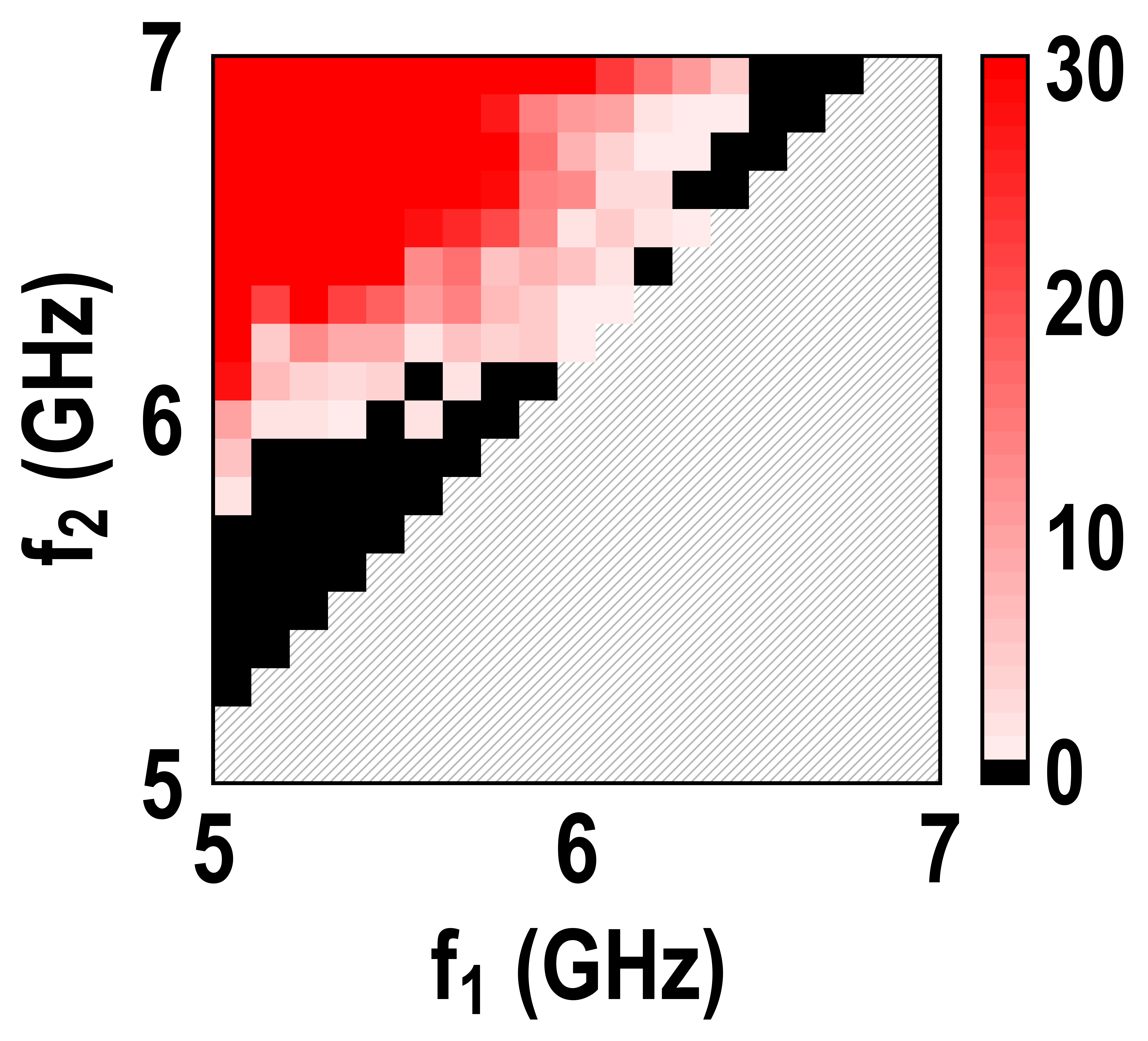}
                &
                \includegraphics[height=0.103\textheight,keepaspectratio]{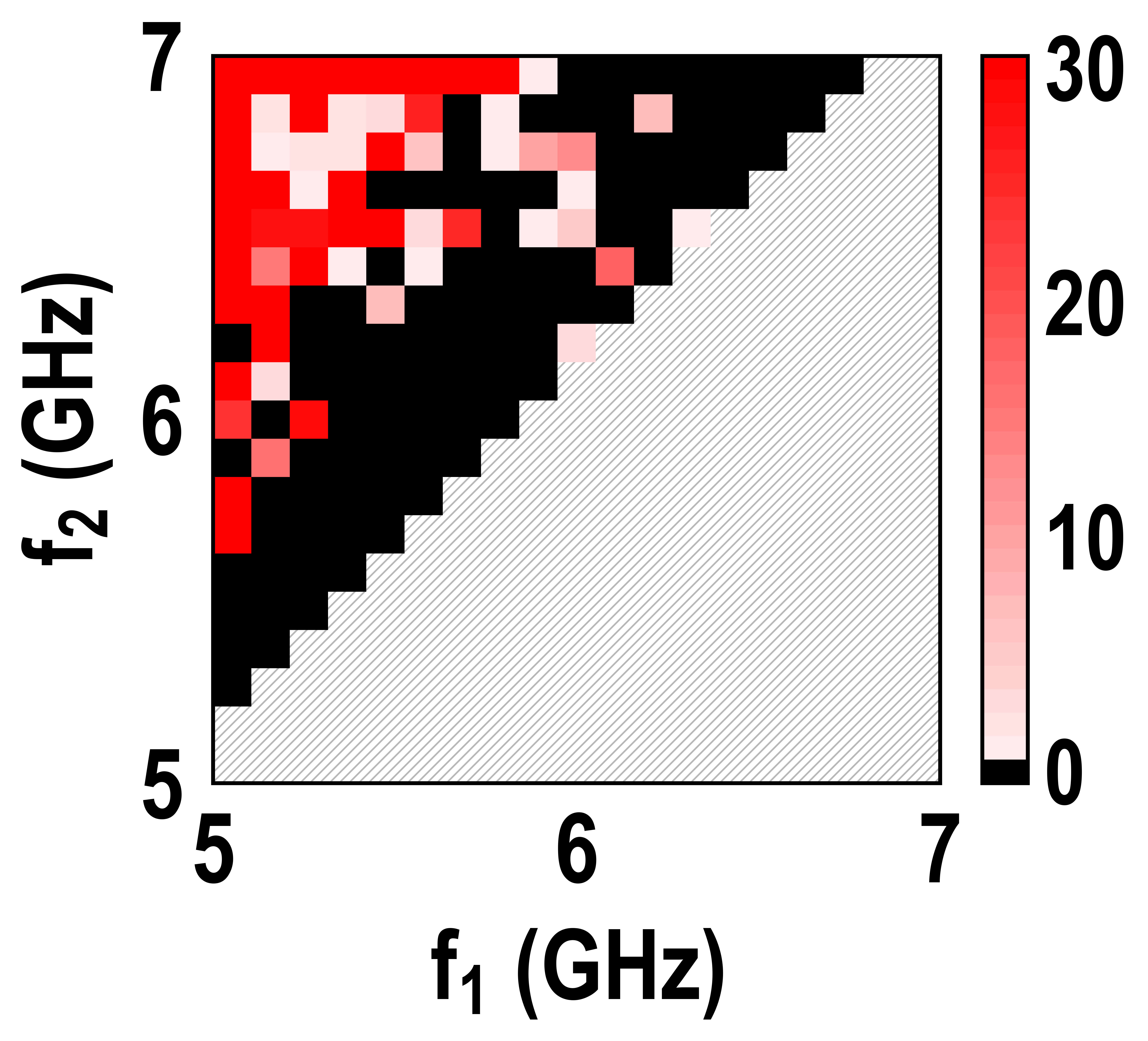}
                &
                \includegraphics[height=0.103\textheight,keepaspectratio]{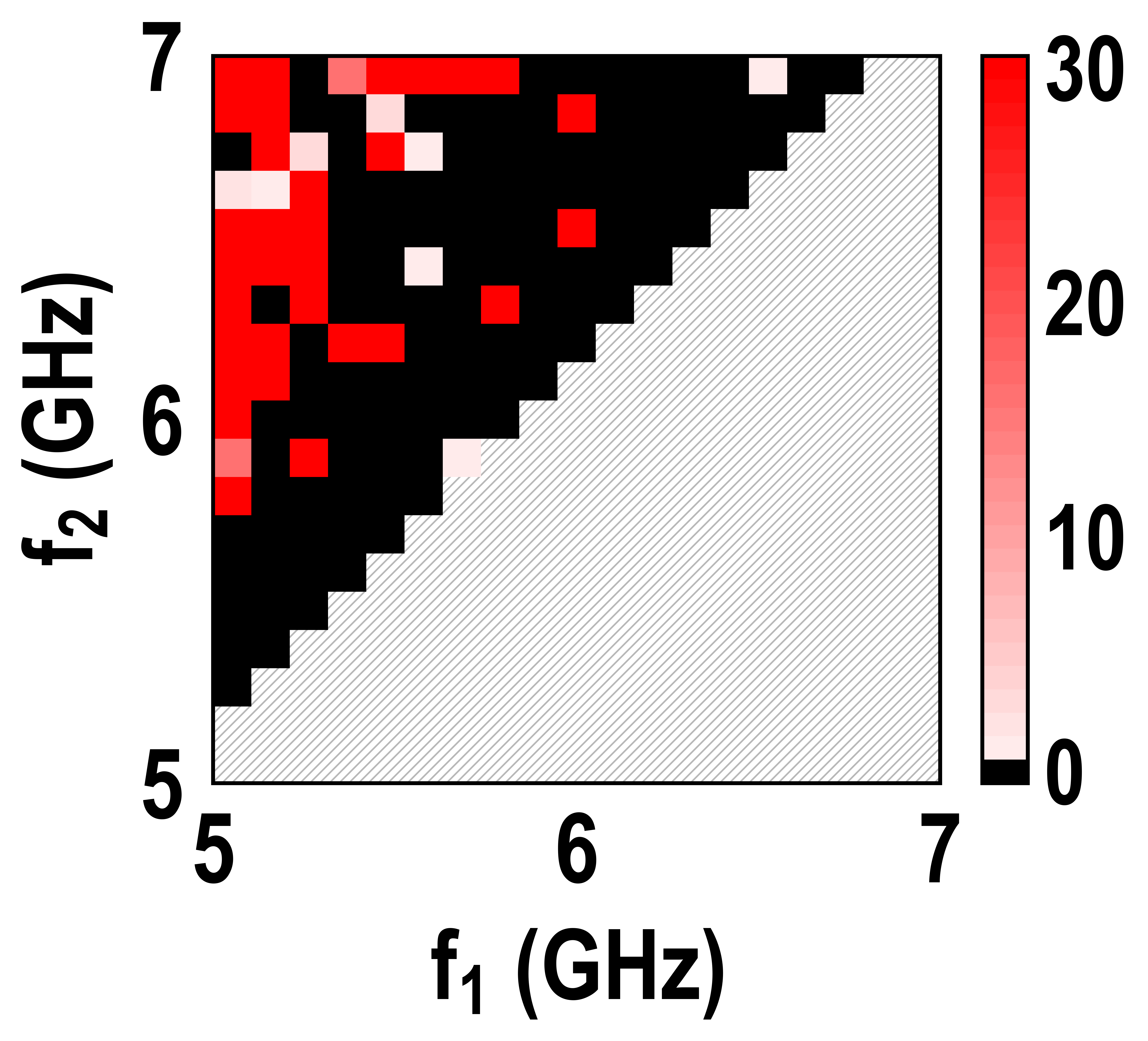}
                \\
                (a) & (b) & (c)
            \end{tabular}}}
    \caption{Heatmaps of unique valid-candidate counts for step-function targets: (a) fine-tuned AR model, (b) BPSO best, and (c) BPSO worst.}
    \label{fig:step_target_heatmaps}
\end{figure}

\begin{table}[!t]
    \centering
    \caption{Comparison of the fine-tuned AR model and BPSO variants}
    \label{tab:step_bpso_coverage}
    \scriptsize
    \color{black}
    \begin{tblr}{
        width = 0.98\columnwidth,
        colspec = {
            |X[1.10,c,m]
            |X[0.90,c,m]
            |X[0.78,c,m]
            |X[1.42,c,m]
            |X[1.25,c,m]|
        },
        hlines,
        rows = {halign=c, valign=m},
        row{1} = {font=\bfseries, ht=5.6ex},
        row{2} = {ht=3.4ex},
        row{3} = {ht=3.4ex},
        row{4} = {ht=3.4ex},
        row{5} = {ht=3.4ex},
        row{6} = {ht=3.4ex},
        rowsep = 0pt,
        colsep = 2pt
    }
        Method
        & {Successful\\cases}
        & {Coverage\\(\%)}
        & {Redundant valid\\candidates (\%)}
        & {Average best\\notch MSE} \\

        Fine-tuned AR
        & 111/153
        & 72.5
        & 0.0
        & 40.15 \\

        BPSO (best)
        & 59/153
        & 38.6
        & 3.5
        & 55.66 \\

        BPSO (worst)
        & 41/153
        & 26.8
        & 37.0
        & 48.31 \\

        VBPSO
        & 43/153
        & 28.1
        & 27.0
        & 59.78 \\

        MVBPSO
        & 61/153
        & 39.9
        & 74.6
        & 52.91 \\
    \end{tblr}
\end{table}

As summarized in Fig.~\ref{fig:step_target_heatmaps} and Table~\ref{tab:step_bpso_coverage}, the fine-tuned AR model achieves substantially broader target-space coverage than conventional BPSO, VBPSO, and MVBPSO while producing non-redundant valid solutions. It also provides the lowest average best notch-region MSE among the evaluated methods, demonstrating more accurate and diverse multi-solution generation for simplified step-function specifications.

\begin{figure}[t]
    \centering
    \setlength{\fboxsep}{1pt}
    \setlength{\fboxrule}{0.8pt}
    \fcolorbox{white}{white}{\includegraphics[width=0.93\linewidth]{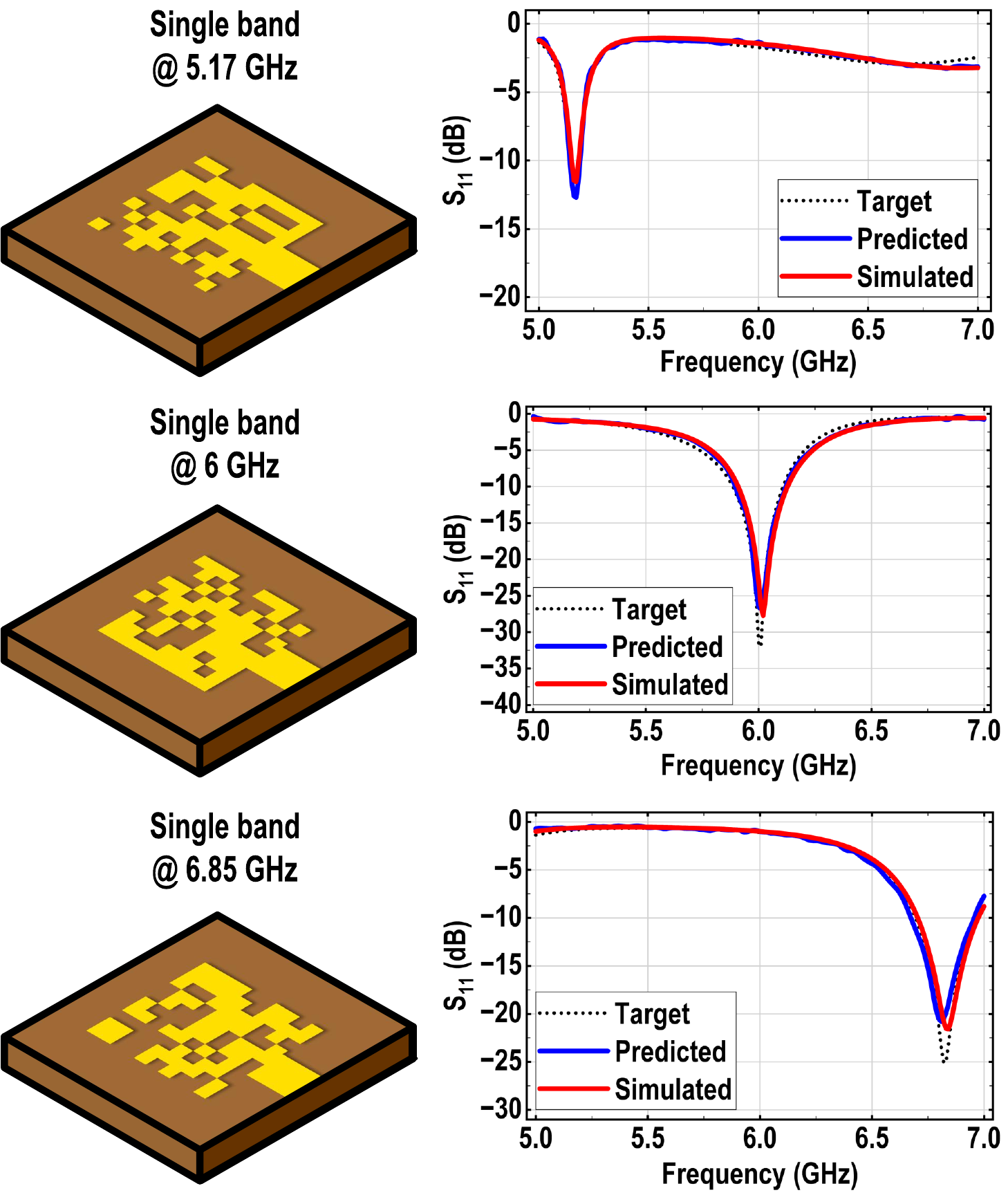}
    }
    \caption{Single-band antenna responses generated by the pretrained AR model.}
    \label{fig:comparison}
\end{figure}

\begin{figure}[t]
    \centering
    \setlength{\fboxsep}{1pt}
    \setlength{\fboxrule}{0.8pt}
    \fcolorbox{white}{white}{\includegraphics[width=0.93\linewidth]{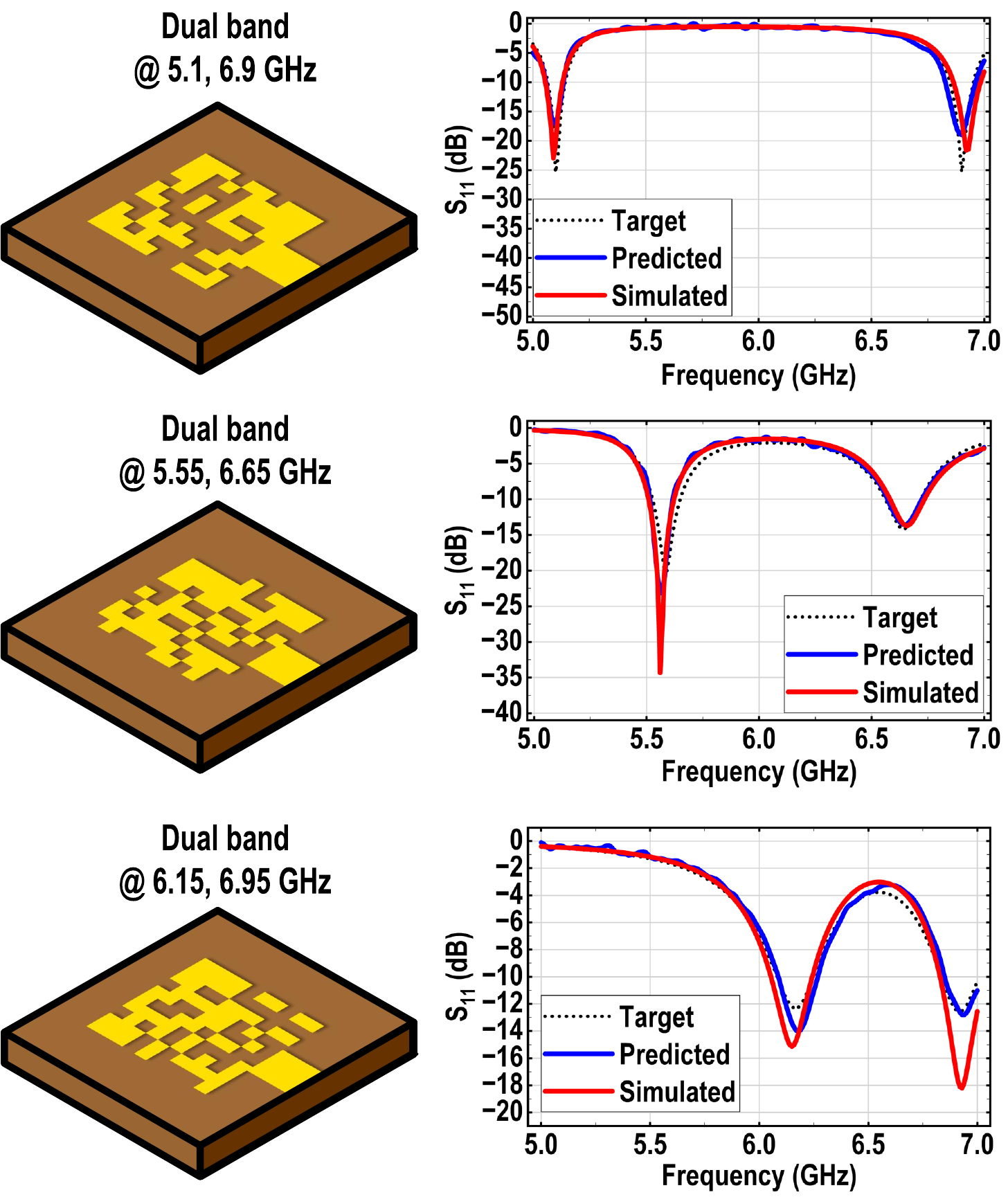}
    }
    \caption{Dual-band antenna responses generated by the pretrained AR model.}
    \label{fig:pretrained_dual}
\end{figure}

\begin{figure}[t]
    \centering
    \setlength{\fboxsep}{1pt}
    \setlength{\fboxrule}{0.8pt}
    \fcolorbox{white}{white}{\includegraphics[width=0.92\linewidth]{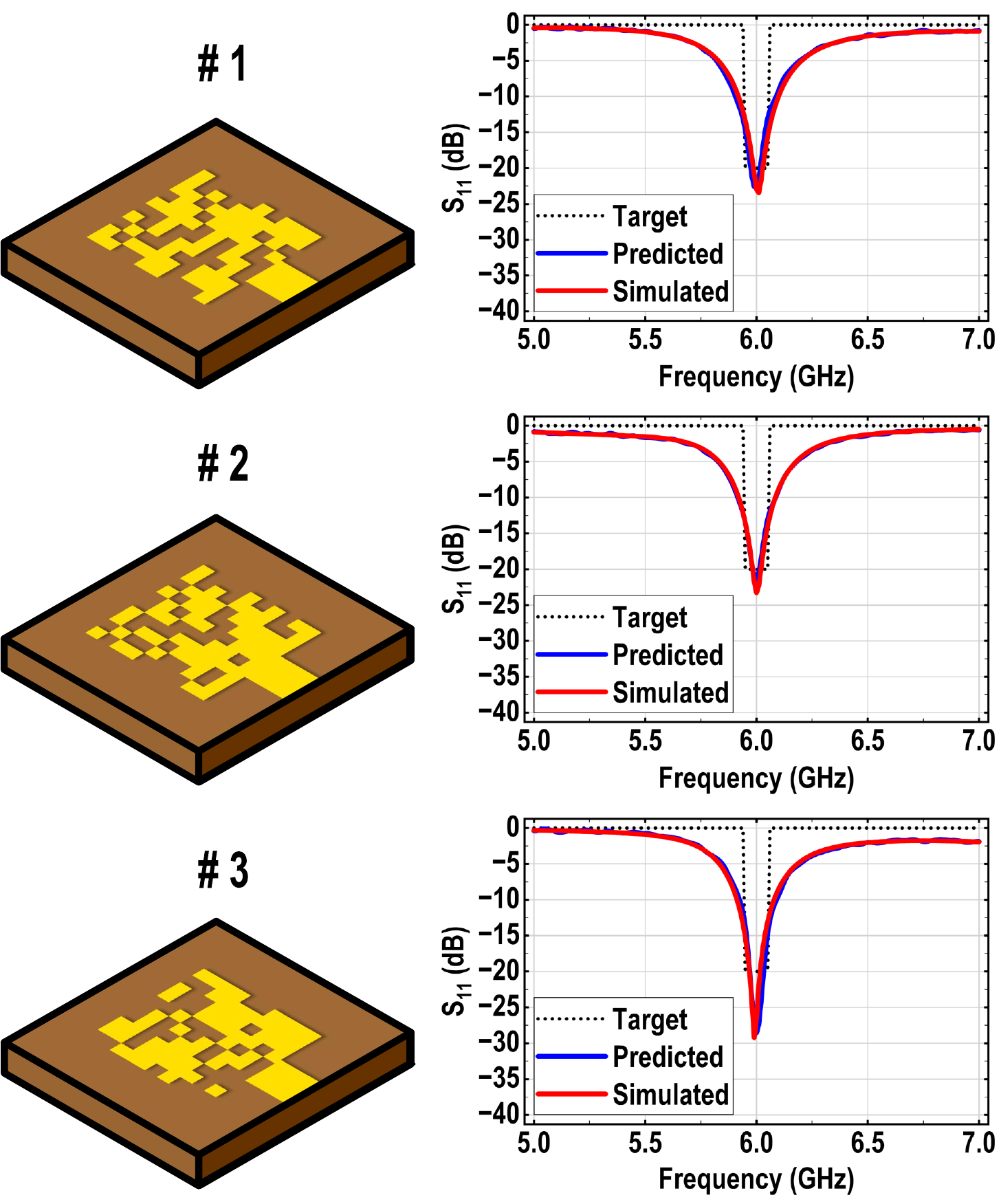}
    }
    \caption{Multiple solutions for a single-band step-function target.}
    \label{fig:multiple}
\end{figure}

\begin{figure}[t]
    \centering
    \setlength{\fboxsep}{1pt}
    \setlength{\fboxrule}{0.8pt}
    \fcolorbox{white}{white}{\includegraphics[width=0.92\linewidth]{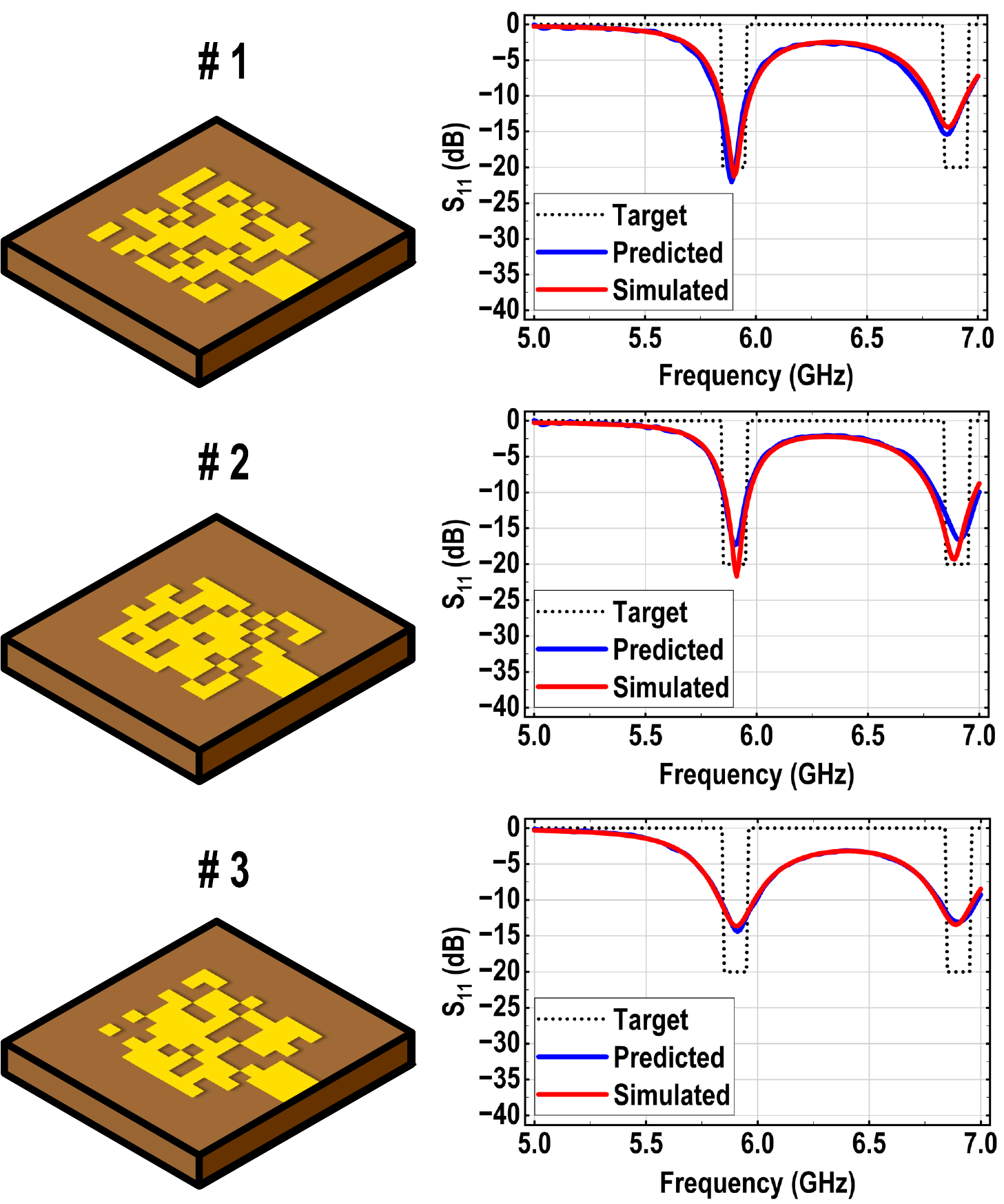}
    }
    \caption{Multiple solutions for a dual-band step-function target.}
    \label{fig:dual_band}
\end{figure}

\subsection{Prediction and Simulation Comparison}

The generation capability of the pretrained AR transformer is first evaluated using single- and dual-band target spectra. Three representative single-band and three representative dual-band test cases are evaluated, as shown in Figs.~\ref{fig:comparison} and~\ref{fig:pretrained_dual}, respectively. The generated layouts exhibit resonant notches close to the target frequencies, while the predicted spectra show good agreement with the full-wave EM simulation results.

To further verify the one-to-many generation capability, the fine-tuned AR transformer is evaluated using step-function targets. The single-band target is specified at 6~GHz, while the dual-band is specified at 5.9 and 6.9~GHz. For each selected target, three distinct antenna candidates are generated, as shown in Figs.~\ref{fig:multiple} and~\ref{fig:dual_band}. Although the generated layouts have different pixel configurations, they produce similar responses corresponding to the same target specification. These results demonstrate that the fine-tuned model can generate multiple distinct solutions for both single- and dual-band step targets.

\begin{figure}[!t]
    \centering
    \includegraphics[width=\linewidth]{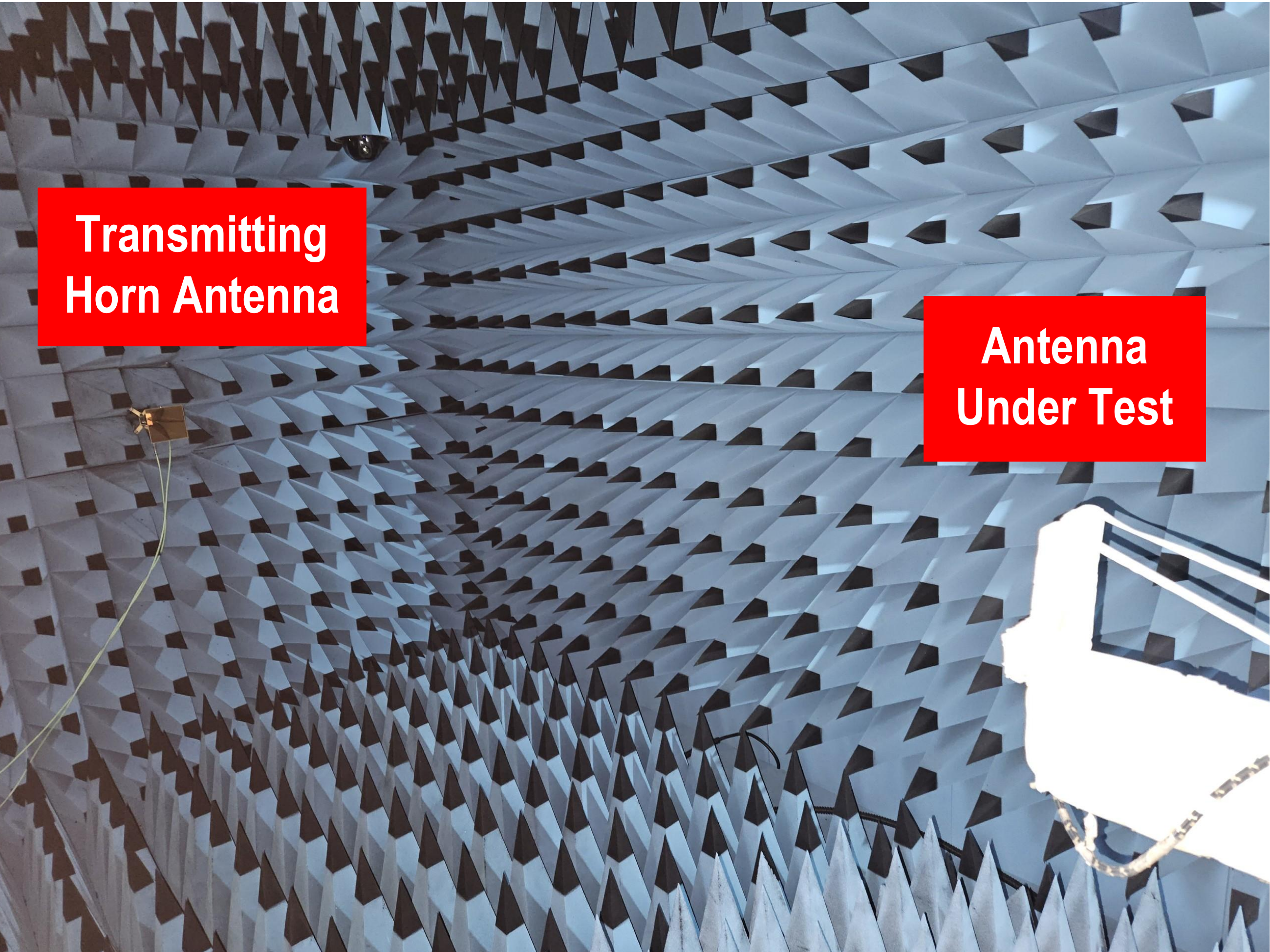}
    \caption{Anechoic chamber setup for radiation pattern measurement.}
    \label{fig:measurement_setup}
\end{figure}

\section{Measurement Results}\label{sec:measurement}

\begin{figure}[!t]
    \centering
    \setlength{\fboxsep}{1pt}
    \setlength{\fboxrule}{0.8pt}
    \fcolorbox{white}{white}{\includegraphics[width=0.98\linewidth]{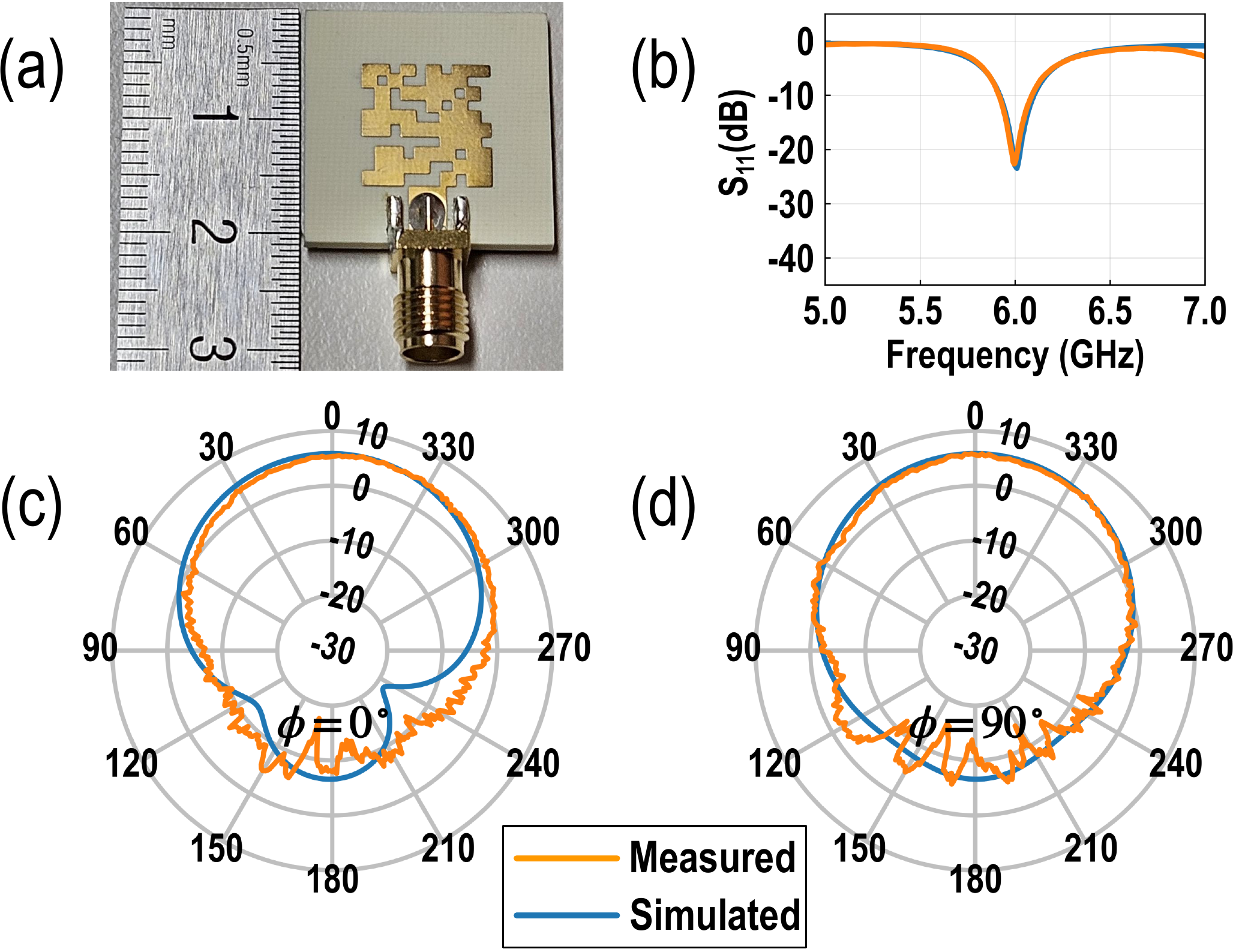}
    }
    \caption{Measured results of the first 6-GHz single-band antenna: 
    (a) fabricated antenna, 
    (b) $S_{11}$ response, 
    (c) Elevation radiation pattern ($\phi = 0^\circ$) at 6.01~GHz,
    (d) Elevation radiation pattern ($\phi = 90^\circ$) at 6.01~GHz.}
    \label{fig:single_6_1}
\end{figure}

\begin{figure}[!t]
    \centering
    \setlength{\fboxsep}{1pt}
    \setlength{\fboxrule}{0.8pt}
    \fcolorbox{white}{white}{\includegraphics[width=0.98\linewidth]{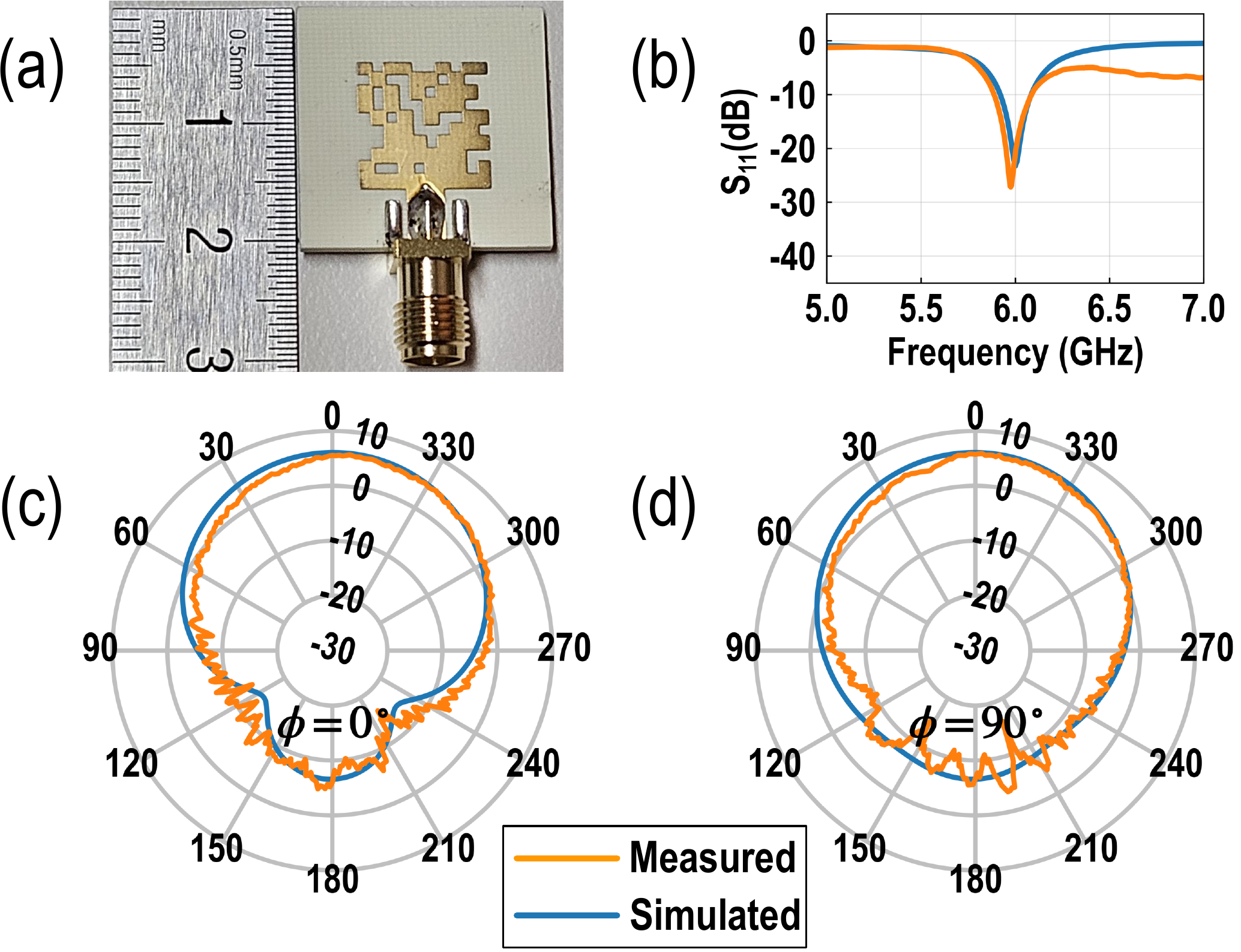}
    }
    \caption{Measured results of the second 6-GHz single-band antenna: 
    (a) fabricated antenna, 
    (b) $S_{11}$ response, 
    (c) Elevation radiation pattern ($\phi = 0^\circ$) at 5.99~GHz,
    (d) Elevation radiation pattern ($\phi = 90^\circ$) at 5.99~GHz.}
    \label{fig:single_6_2}
\end{figure}

\begin{figure}[!t]
    \centering
    \setlength{\fboxsep}{1pt}
    \setlength{\fboxrule}{0.8pt}
    \fcolorbox{white}{white}{\includegraphics[width=0.98\linewidth]{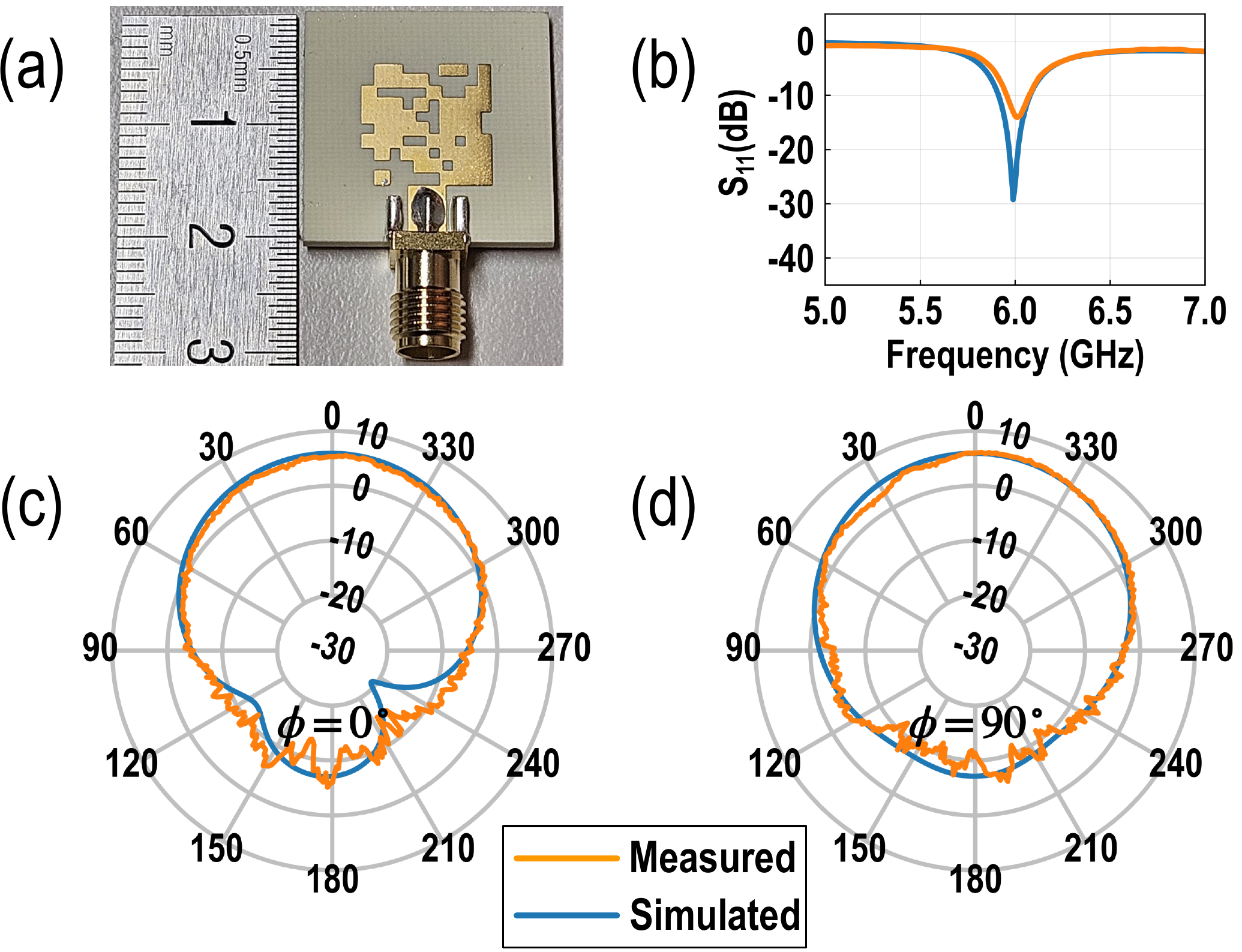}
    }
    \caption{Measured results of the third 6-GHz single-band antenna: 
    (a) fabricated antenna, 
    (b) $S_{11}$ response, 
    (c) Elevation radiation pattern ($\phi = 0^\circ$) at 6.02~GHz,
    (d) Elevation radiation pattern ($\phi = 90^\circ$) at 6.02~GHz.}
    \label{fig:single_6_3}
\end{figure}

To experimentally validate the antenna designs, the antennas are fabricated according to the optimized geometries obtained from the HFSS simulation setup. 
The reflection coefficients are measured using an Anritsu MS46522B 40~GHz vector network analyzer (VNA). 
The radiation characteristics are measured in an anechoic chamber using the measurement setup shown in Fig.~\ref{fig:measurement_setup}. 
In this setup, a transmitting horn antenna is used as the reference source, while the fabricated antenna under test (AUT) is placed at the receiving position. 

Figs.~\ref{fig:single_6_1}--\ref{fig:single_6_3} show the measured characteristics of the three fabricated 6-GHz single-band antennas. 
The fabrication and measurement of these antennas are conducted to experimentally verify the multiple single-band solutions obtained from the simulation results in Fig.~\ref{fig:multiple}. 
The measured $S_{11}$ responses exhibit resonances near 6~GHz and show good agreement with the HFSS simulation results. 
The measured radiation patterns at $\phi = 0^\circ$ and $\phi = 90^\circ$ also follow the simulated results and exhibit broadside radiation characteristics. 
The measured resonant frequency $f_r$, bandwidth BW, and realized gain of each design are summarized in Table~\ref{tab:single_band_measurement}, confirming that all three antennas maintain consistent single-band performance around the target frequency.

\begin{table}[!t]
    \centering
    \caption{Measured performance of 6-GHz single-band antennas}
    \label{tab:single_band_measurement}
    \renewcommand{\arraystretch}{1.2}
    \setlength{\tabcolsep}{4pt}
    \scriptsize
    \resizebox{\columnwidth}{!}{\begin{tabular}{|c|c|c|c|c|}
        \hline
        \multirow{2}{*}{\textbf{Design}} 
        & \multirow{2}{*}{\textbf{$f_r$ (GHz)}} 
        & \multirow{2}{*}{\textbf{BW (MHz)}}
        & \multicolumn{2}{c|}{\textbf{Realized Gain (dBi)}} \\
        \cline{4-5}
        & & & \textbf{$\phi = 0^\circ$} & \textbf{$\phi = 90^\circ$} \\
        \hline
        First  
        & 6.01 
        & 170.20 
        & 5.38 
        & 5.67 \\
        \hline
        Second 
        & 5.99 
        & 192.13 
        & 5.52 
        & 5.74 \\
        \hline
        Third  
        & 6.02 
        & 139.36 
        & 5.32 
        & 5.98 \\
        \hline
    \end{tabular}}
\end{table}

\begin{figure}[!t]
    \centering
    \setlength{\fboxsep}{1pt}
    \setlength{\fboxrule}{0.8pt}
    \fcolorbox{white}{white}{\includegraphics[width=0.97\linewidth]{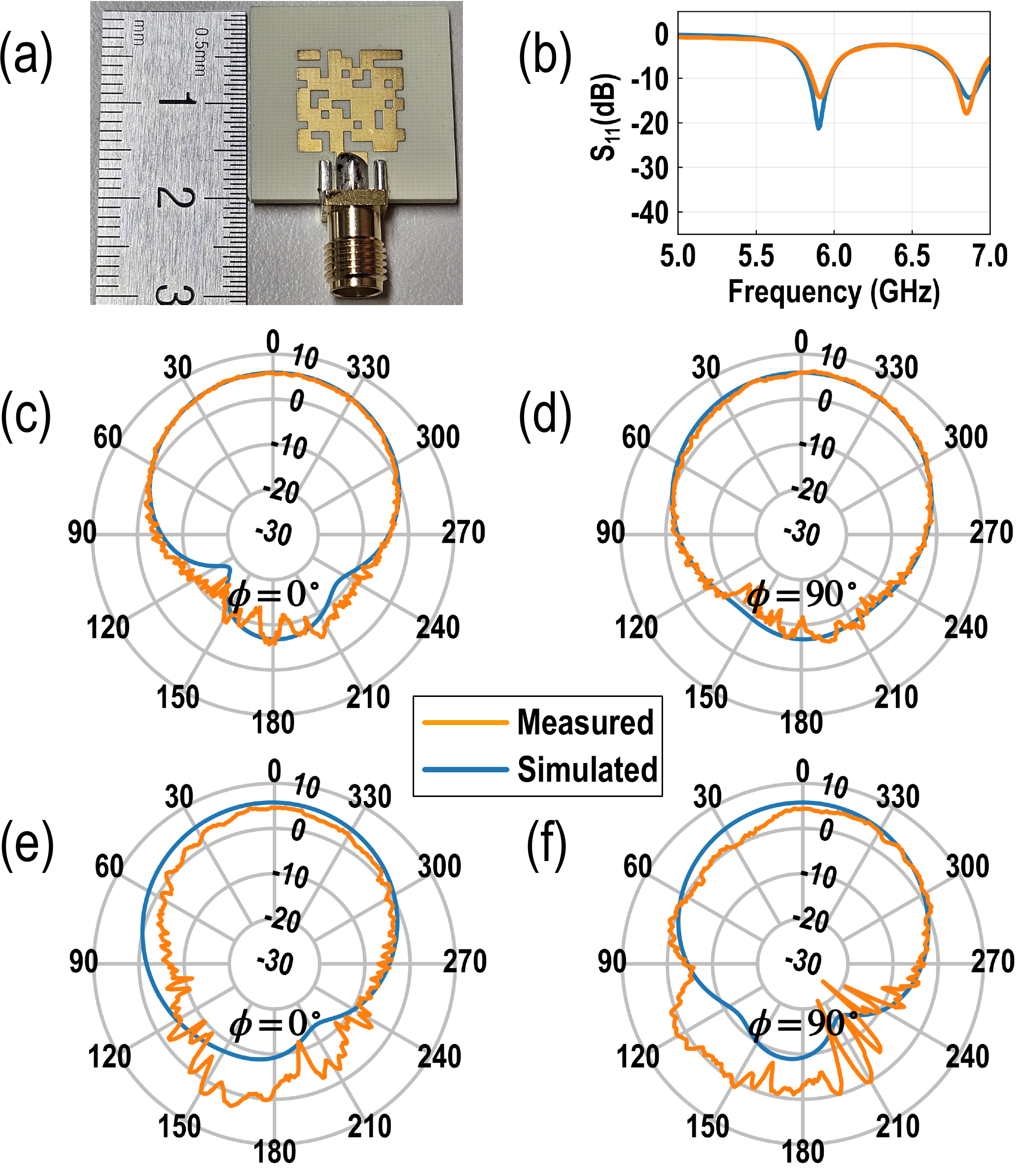}
    }
    \caption{Measured results of the first dual-band antenna: 
    (a) fabricated antenna, 
    (b) $S_{11}$ response, 
    Elevation radiation patterns at $f_1 = 5.91$~GHz 
    (c) $\phi = 0^\circ$, (d) $\phi = 90^\circ$, 
    and at $f_2 = 6.85$~GHz 
    (e) $\phi = 0^\circ$, (f) $\phi = 90^\circ$.}
    \label{fig:dual_1}
\end{figure}

\begin{figure}[!t]
    \centering
    \setlength{\fboxsep}{1pt}
    \setlength{\fboxrule}{0.8pt}
    \fcolorbox{white}{white}{\includegraphics[width=0.97\linewidth]{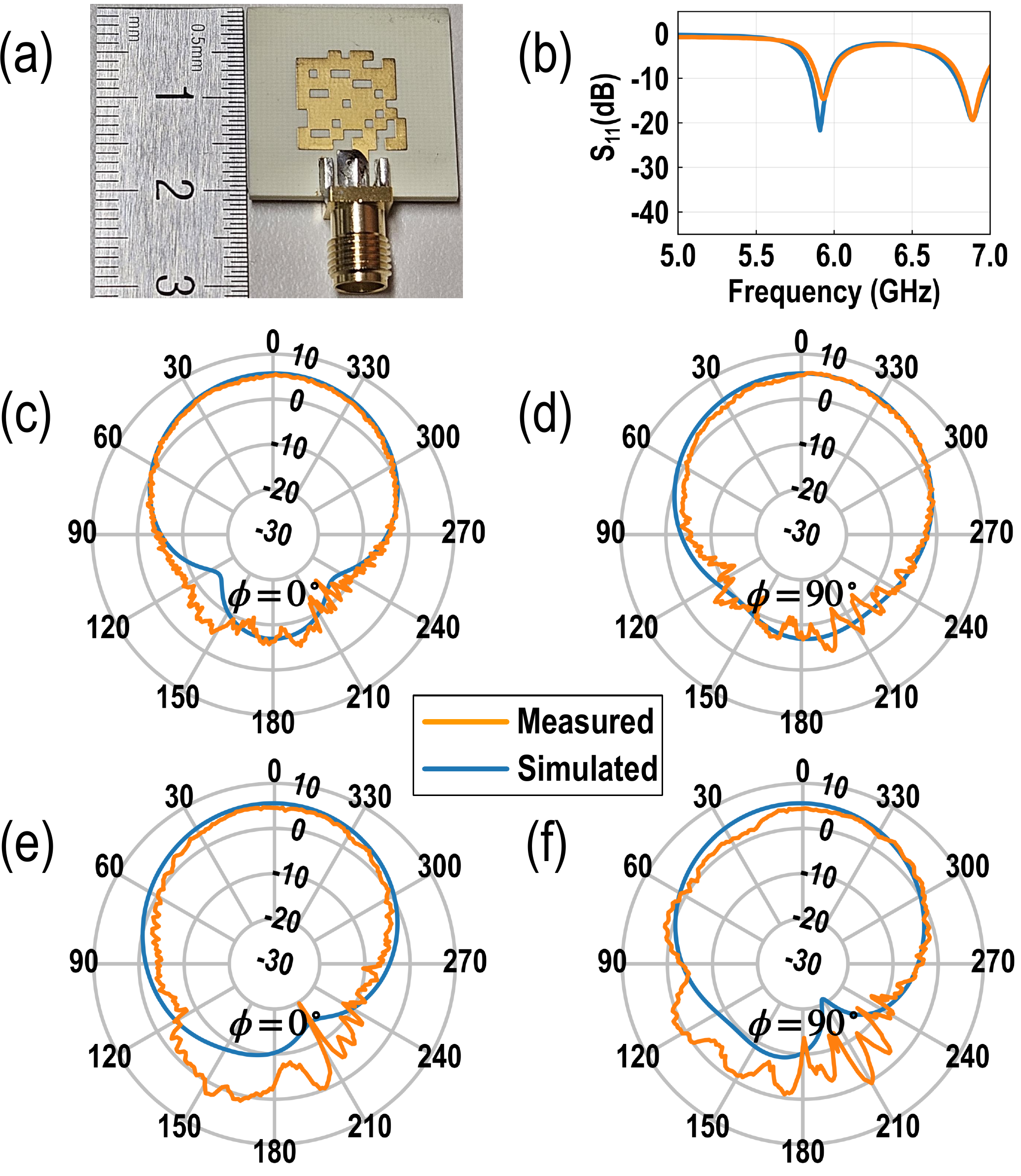}
    }
    \caption{Measured results of the second dual-band antenna: 
    (a) fabricated antenna, 
    (b) $S_{11}$ response, 
    Elevation radiation patterns at $f_1 = 5.94$~GHz 
    (c) $\phi = 0^\circ$, (d) $\phi = 90^\circ$, 
    and at $f_2 = 6.88$~GHz 
    (e) $\phi = 0^\circ$, (f) $\phi = 90^\circ$.}
    \label{fig:dual_2}
\end{figure}

\begin{figure}[!t]
    \centering
    \setlength{\fboxsep}{1pt}
    \setlength{\fboxrule}{0.8pt}
    \fcolorbox{white}{white}{\includegraphics[width=0.97\linewidth]{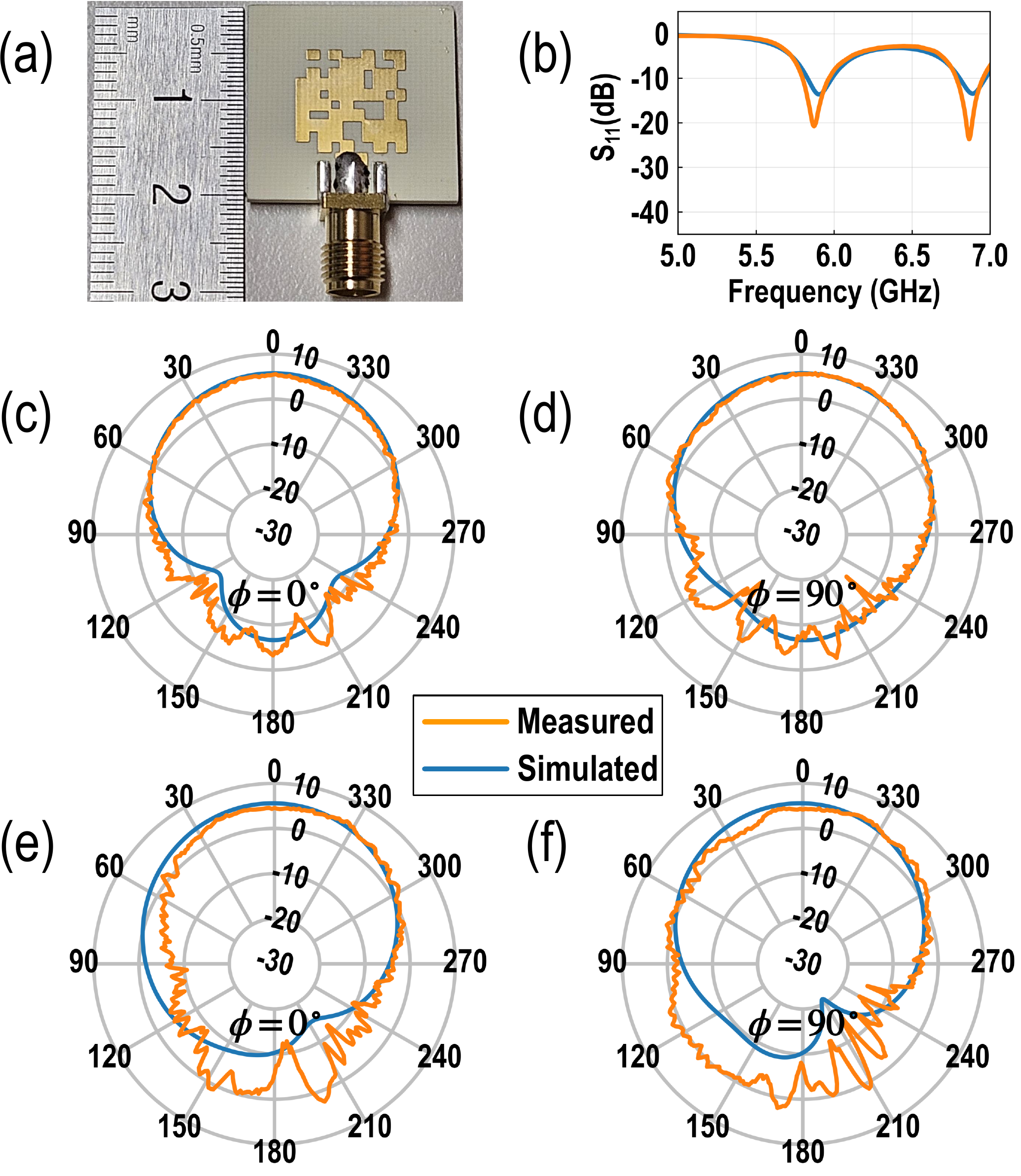}
    }
    \caption{Measured results of the third dual-band antenna: 
    (a) fabricated antenna, 
    (b) $S_{11}$ response, 
    Elevation radiation patterns at $f_1 = 5.87$~GHz 
    (c) $\phi = 0^\circ$, (d) $\phi = 90^\circ$, 
    and at $f_2 = 6.86$~GHz 
    (e) $\phi = 0^\circ$, (f) $\phi = 90^\circ$.}
    \label{fig:dual_3}
\end{figure}

\begin{table}[!t]
    \centering
    \caption{Measured performance of dual-band antennas}
    \label{tab:dual_band_measurement}
    \renewcommand{\arraystretch}{1.3}
    \setlength{\tabcolsep}{4pt}
    \scriptsize
    \resizebox{\columnwidth}{!}{\begin{tabular}{|c|c|c|c|c|c|}
        \hline
        \multirow{2}{*}{\textbf{Design}} 
        & \multirow{2}{*}{\textbf{Band}}
        & \multirow{2}{*}{\textbf{$f_r$ (GHz)}} 
        & \multirow{2}{*}{\textbf{BW (MHz)}}
        & \multicolumn{2}{c|}{\textbf{Realized Gain (dBi)}} \\
        \cline{5-6}
        & & & & \textbf{$\phi = 0^\circ$} & \textbf{$\phi = 90^\circ$} \\
        \hline
        \multirow{2}{*}{First}  
        & Lower 
        & 5.91 
        & 106.85  
        & 5.81 
        & 5.91 \\
        \cline{2-6}
        & Upper 
        & 6.85 
        & 152.08 
        & 4.46 
        & 4.35 \\
        \hline
        \multirow{2}{*}{Second}  
        & Lower 
        & 5.94 
        & 89.53  
        & 5.26 
        & 5.33 \\
        \cline{2-6}
        & Upper 
        & 6.88 
        & 169.03 
        & 4.54 
        & 4.36 \\
        \hline
        \multirow{2}{*}{Third}  
        & Lower 
        & 5.87 
        & 173.76  
        & 5.43 
        & 5.50 \\
        \cline{2-6}
        & Upper 
        & 6.86 
        & 163.68 
        & 4.53 
        & 4.18 \\
        \hline
    \end{tabular}}
\end{table}

To experimentally validate the dual-band solutions obtained from the fine-tuned AR model, three generated dual-band antennas shown in Fig.~\ref{fig:dual_band} are fabricated and measured. 
Figs.~\ref{fig:dual_1}--\ref{fig:dual_3} show the measured characteristics of the first, second, and third generated dual-band antennas, respectively. 
The measured $S_{11}$ responses exhibit two impedance-matching notches near the intended operating bands and show good agreement with the HFSS simulation results. 
The measured radiation patterns generally follow the simulated results and exhibit broadside radiation characteristics, although increased back radiation is observed in the higher-frequency resonance. 
This discrepancy can be attributed in part to diffraction from the finite ground-plane edges, which is known to affect the radiation characteristics of microstrip patch antennas\hypertarget{citeback:huang1983finite}{}\cite{huang1983finite}.
The measured resonant frequency $f_r$, bandwidth BW, and realized gain of each dual-band design are summarized in Table~\ref{tab:dual_band_measurement}.

\section{Discussion}\label{sec:discussion}

\subsection{Quantitative Diversity of Generated Designs}

In addition to the number of unique valid candidates and the redundant-candidate ratio, the structural diversity of the generated layouts is quantitatively evaluated using the normalized Hamming distance. The redundant-candidate ratio identifies exact repetitions, whereas the Hamming distance measures how two nonidentical pixel layouts differ from each other\hypertarget{citeback:wineberg2003underlying}{}\cite{wineberg2003underlying}. For two binary layouts $\mathbf{x}^{(i)}$ and $\mathbf{x}^{(j)}$ containing $P=100$ pixels, the normalized Hamming distance is defined as

\begin{equation*}
\color{black}
d_{\mathrm{H}}\left(\mathbf{x}^{(i)},\mathbf{x}^{(j)}\right)
=
\frac{1}{P}
\sum_{p=1}^{P}
\mathbf{1}
\left[
x_{p}^{(i)} \neq x_{p}^{(j)}
\right],
\qquad P=100.
\end{equation*}

For each target case, exact duplicate layouts are first removed, and the normalized Hamming distance is calculated for every unordered pair of unique valid candidates. The mean distance is obtained for each case containing at least two unique valid candidates and is then averaged across such cases, preventing cases with a large number of generated candidates from dominating the dataset-level metric.

\begin{table}[!t]
    \centering
    \caption{Structural diversity based on normalized Hamming distance}
    \label{tab:hamming_diversity}
    \renewcommand{\arraystretch}{1.3}
    \setlength{\tabcolsep}{4pt}
    \scriptsize
    \resizebox{\columnwidth}{!}{{\color{black}
    \begin{tabular}{|c|c|c|}
        \hline
        \textbf{Target set}
        & \textbf{Method}
        & \textbf{Mean normalized Hamming distance} \\
        \hline
        \multirow{5}{*}{Original targets}
        & Pretrained AR & 0.4540 \\
        \cline{2-3}
        & BPSO (best) & 0.1351 \\
        \cline{2-3}
        & BPSO (worst) & 0.1174 \\
        \cline{2-3}
        & VBPSO & 0.2680 \\
        \cline{2-3}
        & MVBPSO & 0.1678 \\
        \hline
        \multirow{5}{*}{Step-shaped targets}
        & Fine-tuned AR & 0.4605 \\
        \cline{2-3}
        & BPSO (best) & 0.1606 \\
        \cline{2-3}
        & BPSO (worst) & 0.0895 \\
        \cline{2-3}
        & VBPSO & 0.2159 \\
        \cline{2-3}
        & MVBPSO & 0.1156 \\
        \hline
    \end{tabular}}}
\end{table}

As shown in Table~\ref{tab:hamming_diversity}, both the pretrained and fine-tuned AR models achieve substantially higher normalized Hamming distances than conventional BPSO, VBPSO, and MVBPSO. Together with the low redundant-candidate ratios, this indicates that the proposed AR framework generates valid antenna layouts with greater structural diversity and less repetition across both original and step-shaped target sets.

\subsection{Model Scaling and Computational Complexity}

The present framework uses a $10\times10$ binary pixel grid, corresponding to an autoregressive sequence of 100 pixels. For an $M\times M$ grid, the sequence length becomes $N=M^2$, and the self-attention operation scales quadratically with sequence length. Thus, increasing the resolution to $20\times20$ or $30\times30$ increases the number of attention interactions by 16 and 81 times, respectively. Although a finer grid provides additional geometric degrees of freedom, the heatmap results over unseen dual-band targets show that the current $10\times10$ representation already achieves high target-space coverage. Even within this $10\times10$ design space, the BPSO variants have difficulty covering a wide range of target specifications, whereas the AR transformer maintains substantially broader coverage.

Increasing the grid resolution does not necessarily improve the performance of the AR transformer, because a larger grid expands the discrete design space and makes the inverse mapping more difficult. Therefore, simply increasing the grid size does not guarantee higher generation success or convergence to a valid solution. The $10\times10$ representation may instead provide a practical balance between sufficient geometric flexibility and manageable computational complexity for the dual-band design problem considered in this work.

\subsection{Hyperparameter Selection and Model Tuning}

The architectural hyperparameters of both the forward surrogate and the AR inverse model were determined through systematic hyperparameter searches rather than being selected arbitrarily. Weights \& Biases\hypertarget{citeback:wandb}{}\cite{wandb} sweeps with Bayesian optimization were used to explore candidate configurations while continuously monitoring the corresponding validation performance. For the forward surrogate, this process included the number and arrangement of residual blocks and channel dimensions. For the AR transformer, major parameters including the number of Transformer layers, hidden dimension, attention heads, feed-forward dimension, dropout ratio, and learning rate were examined. The final configuration was selected based on the best validation behavior observed during these sweeps.

\subsection{Effect of the Fixed Pixel Density}

Another design choice in the present framework is the use of a fixed 60\% metallic-pixel density. The density was selected based on the preliminary statistical analysis, which showed that this region provides the highest probability of obtaining resonant antenna responses. Concentrating the training data on this resonance-rich region reduces the number of non-resonant samples presented to the model and allows the available training budget to be used more efficiently for learning useful antenna geometries.

However, fixing the density also restricts the accessible design space~\cite{khaireh2023newcomer}. Low-density or extremely dense layouts may exhibit electromagnetic behavior that is not represented by the present model, and the current results should therefore not be interpreted as demonstrating optimality over all possible pixel densities. Extending the training set to multiple density levels would provide a wider design space, although this would also increase the amount of training data and model capacity. Such an extension is particularly attractive for future multi-objective inverse design, where $S_{11}$ can be optimized together with additional objectives such as gain, radiation efficiency, polarization, or bandwidth rather than using the reflection response alone.

\subsection{Comparison with Diffusion-Based Models}

Diffusion models have recently been widely investigated for generative inverse-design problems because of their ability to represent complex multimodal design distributions\hypertarget{citeback:kim2026mr2p}{}\cite{kim2026mr2p},\hypertarget{citeback:zhang2025synthesis}{}\cite{zhang2025synthesis}. However, the present antenna geometry consists of only a $10\times10$ binary pixel grid, for which an autoregressive representation provides a more direct formulation. Each pixel can be generated sequentially as a discrete binary variable conditioned on the target spectrum and previously generated pixels, naturally enforcing the binary design constraint without an additional continuous-to-binary conversion or iterative denoising process.

In contrast, diffusion-based generation generally requires multiple denoising steps for each candidate, introducing additional sampling complexity that may be unnecessary for this relatively low-dimensional discrete design space. Therefore, the AR architecture was selected as a practical balance between multi-solution generation capability and computational complexity for the $10\times10$ pixelated antenna problem. Diffusion-based approaches remain an interesting direction for future work, particularly for higher-dimensional or more complex geometric representations.

\subsection{Extension to Other Antenna Configurations}

The present model is developed for a fixed antenna configuration, including the substrate and feed geometry. Based on previous studies applying transfer learning to antenna and metasurface inverse design\hypertarget{citeback:fan2022transfer}{}\hypertarget{citeback:kanmaz2023deep}{}\cite{gupta2024transfer,fan2022transfer,kanmaz2023deep}, the proposed framework could potentially be adapted to other substrate properties, feed structures, or antenna dimensions by fine-tuning the pretrained model with a relatively small amount of additional data from the new configuration. By reusing the features and design relationships learned from the original dataset, transfer learning can reduce the amount of newly generated training data and shorten the retraining process compared with training a new model from scratch. Nevertheless, the amount of additional data required is expected to depend on the degree of difference between the original and target configuration.

\subsection{Pixelated-Patch Antenna Design Considerations}

Pixelated antennas provide a large discrete design space and enable compact and multiband geometries that are difficult to obtain using only a small number of conventional geometric parameters. At the same time, irregular and asymmetric metallic distributions introduce additional design considerations. In particular, asymmetry can degrade co- and cross-polarization isolation and therefore reduce polarization purity. Imposing strict geometric symmetry during generation is one possible way to improve polarization purity, but doing so substantially reduces the available pixel design space. Kumari \textit{et al.} demonstrated that the high cross-polarization associated with an asymmetric inverse-designed pixelated antenna can be significantly suppressed by arranging mirror-symmetric elements with differential feeding, improving the broadside cross-polarization level from $-5$ to $-50$~dB\hypertarget{citeback:kumari2025polarization}{}\cite{kumari2025polarization}.

Another practical issue in pixelated antenna design is the electrical connectivity between neighboring conducting pixels. Infinitesimal point contacts that are well defined in ideal EM simulations may be difficult to reproduce reliably because of fabrication tolerances. Previous studies have therefore introduced finite overlaps between adjacent pixels to ensure electrical connectivity\hypertarget{citeback:wu2025efficient}{}\cite{wu2025efficient,chen2026surrogate}. Jayasinghe \textit{et al.} further investigated the effect of different overlap sizes, showing that the overlap configuration can also affect the resulting antenna performance\hypertarget{citeback:jayasinghe2015nonuniform}{}\cite{jayasinghe2015nonuniform}. Accordingly, in this work, a 0.25-mm overlap is introduced between neighboring pixels to ensure reliable electrical connectivity and improve manufacturability.

In addition to the issues arising from pixelated structures, microstrip patch antennas inherently exhibit a relatively narrow impedance bandwidth because of their resonant nature\hypertarget{citeback:lee2012microstrip}{}\cite{lee2012microstrip}. This limitation has traditionally been alleviated through structural and feeding modifications. For example, aperture-coupled stacked patches have been employed to broaden the operating bandwidth through coupled resonances\hypertarget{citeback:targonski1998design}{}\cite{targonski1998design}, while defected ground structures (DGSs) have been used to modify the ground-current distribution and enhance impedance bandwidth\hypertarget{citeback:marotkar2016bandwidth}{}\cite{marotkar2016bandwidth}. More recently, geometry optimization has provided another approach to broadband patch design; Wang \textit{et al.} demonstrated machine-learning-assisted bandwidth optimization of pixelated patch antennas, achieving measured relative bandwidths of 24.5\% and 32.5\%\cite{wang2024machine}. These studies demonstrate that the inherent bandwidth limitation of patch antennas can be substantially mitigated through appropriate structural and geometry optimization.

Table~\ref{tab:dual_band_comparison} compares the proposed method with state-of-the-art dual-band antenna design works in terms of design methodology, target specification capability, synthesis time, and antenna performance. Many prior inverse-design methods either return a single representative geometry for a given target response or do not explicitly support step-function-based target specifications. In contrast, our work directly generates multiple 10~$\times$~10 pixelated patch antenna layouts from a desired dual-band step-function target within approximately one minute. Therefore, this work provides a practical balance among automated inverse design, target-driven synthesis, multiple-solution generation, and experimental feasibility.

\begin{table*}[!t]
    \centering
    \caption{Comparison with state-of-the-art dual-band antenna works in terms of design methodology and antenna performance}
    \label{tab:dual_band_comparison}
    \scriptsize
    \resizebox{0.995\textwidth}{!}{\begin{tblr}{
        colspec = {
            |Q[c,m,0.55cm]
            |Q[c,m,2.4cm]
            |Q[c,m,0.95cm]
            |Q[c,m,0.95cm]
            |Q[c,m,1.55cm]
            |Q[c,m,2.1cm]
            |Q[c,m,1.85cm]
            |Q[c,m,1.25cm]
            |Q[c,m,1.25cm]
            |Q[c,m,1.55cm]|
        },
        hlines,
        rows = {halign=c, valign=m},
        row{1} = {font=\bfseries, ht=9ex},
        row{2} = {ht=8.5ex},
        row{3} = {ht=9.5ex},
        row{4} = {ht=9.5ex},
        row{5} = {ht=8.5ex},
        row{6} = {ht=9.5ex},
        row{7} = {ht=9.5ex},
        row{8} = {ht=9.5ex},
        row{9} = {ht=8.5ex},
        row{10} = {ht=8.5ex},
        row{11} = {ht=9.5ex},
        row{12} = {ht=9.5ex},
        row{13} = {ht=9.5ex},
        row{14} = {ht=9.5ex},
        row{15} = {ht=8.5ex},
        rowsep = 0pt
    }
        Ref. 
        & {Design\\Method}
        & {Multiple\\solutions}
        & {Step-function\\target}
        & {Average\\synthesis time}
        & {Antenna\\type}
        & {Substrate /\\height}
        & {$f_1/f_2$\\(GHz)}
        & {Realized\\gain (dBi)}
        & {Antenna\\efficiency} \\

        {\hypertarget{citeback:liu2023differentially}{}\cite{liu2023differentially}}
        & {Conventional\\heuristic design}
        & X
        & X
        & NR
        & {Coupled-resonator\\patch antenna}
        & {Rogers 4003 /\\11 mm}
        & {3.24--3.83 /\\4.74--5.30}
        & 8.1 / 8.8
        & NR \\

        {\hypertarget{citeback:bui2023dual}{}\cite{bui2023dual}}
        & {Conventional\\heuristic design}
        & X
        & X
        & NR
        & {Differential-fed\\slotted-patch}
        & {Rogers RT \\duroid 5880 /\\7.35 mm}
        & {2.40--2.52 /\\5.09--5.41}
        & 8.5 / 9.2$^{*}$
        & {$\geq$80 / $\geq$80\%$^{*}$} \\

        {\hypertarget{citeback:yang2021frequency}{}\cite{yang2021frequency}}
        & {Conventional\\heuristic design}
        & X
        & X
        & NR
        & {Frequency selective-\\surface based shared aperture patch}
        & {Rogers RO4003 /\\25.6 mm}
        & {2.3--2.7 /\\3.3--3.8}
        & 9.1 / 8.6
        & 85 / 76\% \\

        {\cite{gupta2023tandem}}
        & {Tandem NN\\inverse design}
        & X
        & X
        & $<$1 s
        & {12 $\times$ 12 pixelated\\microstrip antenna}
        & {Air /\\0.61 mm}
        & 14 / 18.37
        & 9.73 / 9.43$^{*}$
        & 92.3 / 91.4\%$^{*}$ \\

        {\cite{yuan2025tandem}}
        & {TGVAE-FGNN\\inverse design}
        & O
        & O
        & {NR}
        & {Multiband slotted\\dipole antenna}
        & {Brass sheet /\\NR}
        & {Target-dependent\\0.3--3.0}
        & NR
        & NR \\

        {\cite{gupta2024transfer}}
        & {Transfer-learning surrogate + BPSO\\ inverse design}
        & O
        & O
        & 40 min
        & {12 $\times$ 12 pixelated\\microstrip antenna}
        & {FR4 /\\3.2 mm}
        & 4.0 / 4.7
        & 4.12 / 4.31
        & NR \\

        {\cite{chen2025inverse}}
        & {CNN surrogate +\\BPSO inverse design}
        & X
        & X
        & 30 min
        & {10 $\times$ 10 pixelated\\microstrip antenna}
        & {Ideal lossless, $\epsilon_r=2.2$ /\\2.5 mm}
        & 2.8 / 3.7
        & 7.07 / 6.85$^{*}$
        & 95.3 / 98.8\%$^{*}$ \\

        {\cite{ullah2023multiservice}}
        & {VBPSO-based\\EM optimization}
        & X
        & X
        & 38.47 h
        & {Triangular- / square-\\pixel stacked patch}
        & {Rogers 4003 /\\3.04 mm}
        & {5.16--5.21 /\\5.79--5.86}
        & 3.82 / 3.46
        & 70 / 82\%$^{*}$ \\

        {\hypertarget{citeback:dou2026multi}{}\cite{dou2026multi}}
        & {SADRL inverse design\\(IBPSO+BCNN$_1$+DQN)}
        & X
        & X
        & 15.43 h
        & {12 $\times$ 10 pixelated\\ monopole antenna}
        & {FR4 /\\1.6 mm}
        & {3.3--3.8 /\\5.75--5.85}
        & {1.89$\pm$0.23 /\\1.35$\pm$0.11}
        & NR \\

        {\cite{wu2025efficient}}
        & {DC-SADEA\\with XGBoost\\inverse design}
        & X
        & X
        & 9.3 days
        & {Digitally coded feeding-crossed\\dipole antenna}
        & {FR4 /\\18.8 mm}
        & {3.3--3.8 /\\4.8--5.0}
        & 6.47 / 5.31$^{*}$
        & NR \\

        {\hypertarget{citeback:papathanasopoulos2023optimization}{}\cite{papathanasopoulos2023optimization}}
        & {ONN-assisted\\EM optimization}
        & X
        & X
        & 4 h
        & {Coaxial probe-fed\\slotted patch}
        & {Rogers RT \\duroid 5880 /\\3.175 mm}
        & 1.9 / 2.4
        & NR
        & NR \\

        {\hypertarget{citeback:yuan2025flex}{}\cite{yuan2025flex}}
        & {FGNN-based\\inverse design}
        & X
        & O
        & NR
        & {Variable-slot\\monopole patch}
        & {Brass sheet /\\NR}
        & {Target-dependent\\0.3--3.0}
        & NR
        & NR \\

        {\cite{pang2025inverse}}
        & {Diffusion model +\\BCNN$_2$-CMA\\inverse design}
        & O
        & X
        & NR
        & {Circularly polarized\\microstrip antenna}
        & NR
        & 3.63 / 6.91
        & 6.61 / 3.12
        & NR \\

        This work
        & {AR transformer\\inverse design}
        & O
        & O
        & $\sim$1 min
        & {10 $\times$ 10 pixelated\\microstrip antenna}
        & {Rogers 4350B /\\1.52 mm}        
        & {5.80--5.97 /\\6.78--6.95}
        & 5.43 / 4.53
        & 88.0 / 87.7\%$^{*}$ \\
    \end{tblr}}
    \vspace{0.1mm}
    \begin{flushleft}
        \scriptsize$^{*}$Simulated value.
        NR: Not reported. NN: Neural network. TGVAE: Tandem generalized variational autoencoder. FGNN: Flex-form generative neural network. VBPSO: V-shaped transfer function based binary particle swarm optimization. SADRL: Surrogate model-assisted deep reinforcement learning. IBPSO: Improved binary particle swarm optimization. BCNN$_1$: Bayesian convolutional neural network. DQN: Deep Q-network. DC-SADEA: Digitally coded antenna-oriented surrogate model-assisted evolutionary algorithm. XGBoost: Extreme gradient boosting. ONN: Optimization assisted by a neural network. BCNN$_2$: Bilinear convolutional neural network. CMA: Characteristic mode analysis.
    \end{flushleft}
\end{table*}

\section{Conclusion}\label{sec:conclusion}

This paper demonstrates a fine-tuned autoregressive transformer for efficient inverse design of pixelated patch antennas. While conventional metaheuristic algorithms such as BPSO require iterative searches and are highly affected by initial conditions and hyperparameter tuning, the proposed AR transformer learns the conditional distribution between target $S_{11}$ responses and antenna layouts, enabling direct generation of valid antenna candidates. The comparison with BPSO shows that the proposed method achieves more efficient valid-candidate generation and provides diverse antenna solutions. Furthermore, step-shaped target spectrum is introduced for fine-tuning, allowing practical dual-band inverse design through simple band-level specifications. The fabricated antennas show resonant responses and broadside radiation characteristics consistent with the intended designs. These results confirm that the proposed AR transformer overcomes key limitations of metaheuristic-based inverse design and offers an efficient framework for multi-solution antenna synthesis.

\section*{Acknowledgement}

The authors would like to thank Prof. Jaeho Im of Hongik University for providing the measurement equipment used in this work.

\bibliographystyle{IEEEtran}
\bibliography{references}

\begin{IEEEbiography}
[{\YMBioPhoto{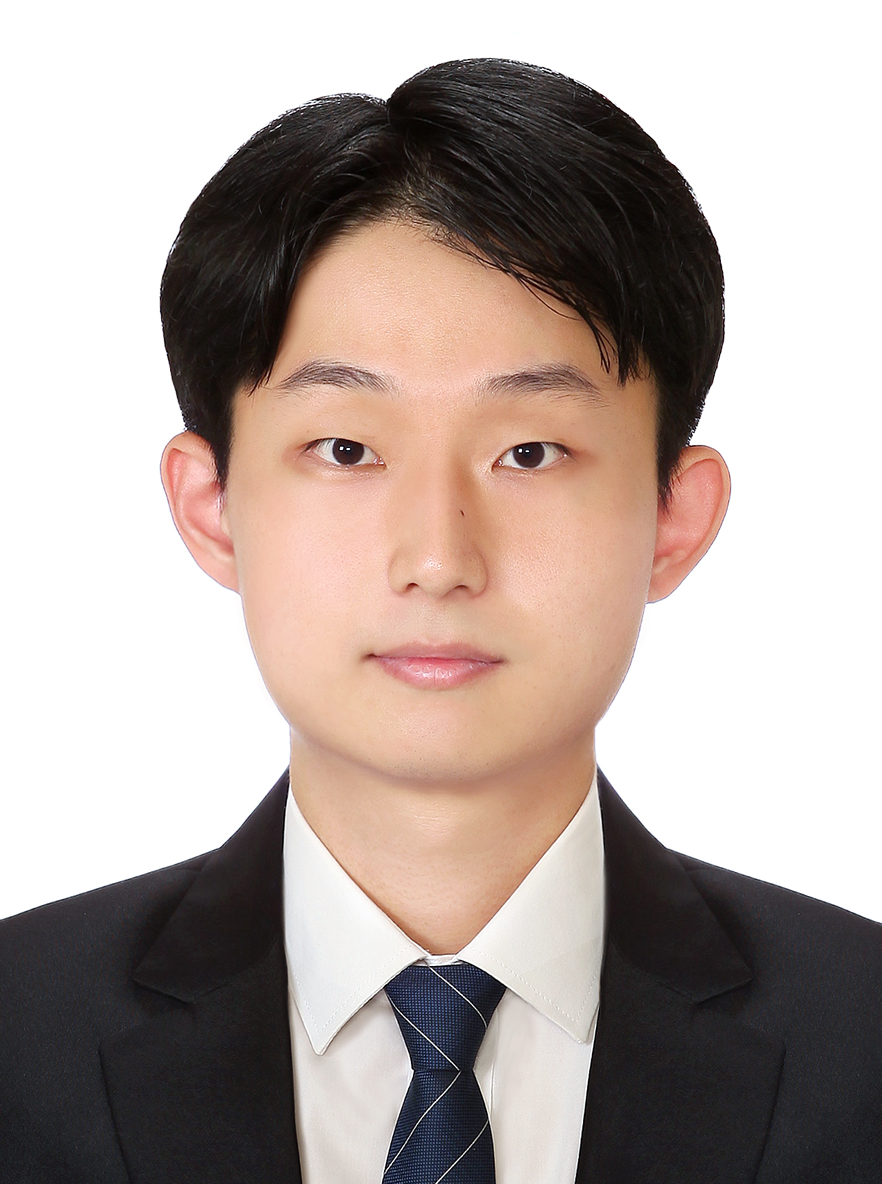}}]
{Young-Mok Yun}
received the B.S. degree in electrical and electronic engineering from Hongik University, Seoul, South Korea, in 2025. He is currently working toward the M.S. degree in system semiconductor engineering at Yonsei University. His current research interests include AI-assisted inverse design and optimization of RF/millimeter-wave components and high-speed interface systems.
\end{IEEEbiography}

\vspace{-2.55\baselineskip}

\begin{IEEEbiography}
[{\YMBioPhoto{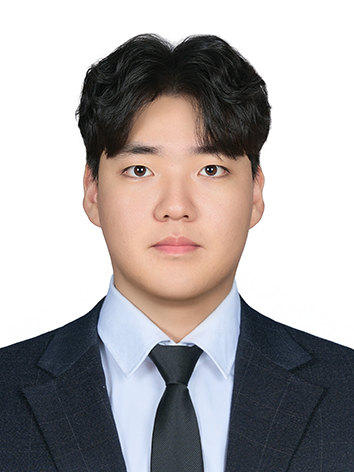}}]
{Seung-Won Shin}
received the B.S. degree in electrical and electronic engineering from Yonsei University, Seoul, South Korea, in 2026. He is currently working toward the M.S. degree in system semiconductor engineering at Yonsei University. His current research interests include AI-assisted integrated circuit design and machine-learning-based electronic design automation.
\end{IEEEbiography}

\vspace{-2.55\baselineskip}

\begin{IEEEbiography}
[{\YMBioPhoto{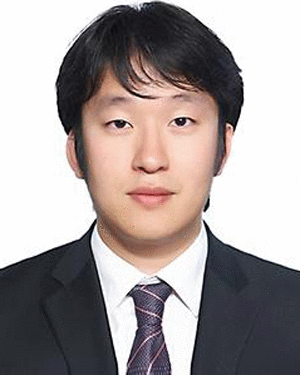}}]
{Jaeyong Chung}
(Member, IEEE) received the B.S. degree in electrical engineering from Yonsei University, Seoul, South Korea, in 2006, and the M.S. and Ph.D. degrees in electrical and computer engineering from the Department of Electrical and Computer Engineering, University of Texas at Austin, Austin, TX, USA, in 2008 and 2011, respectively.

He worked with the Strategic CAD Lab (SCL), Intel, and the IBM Thomas J. Watson Research Center in Summer 2008 and Summer 2010, respectively. From 2011 to 2013, he was with the Design Compiler Team, Synopsys, Inc., Mountain View, CA, USA. He is currently an Associate Professor with Yonsei University, Seoul, South Korea. His current research interests include efficient deep learning and AI-based hardware design.
\end{IEEEbiography}

\vspace{-1.55\baselineskip}

\begin{IEEEbiography}
[{\YMBioPhoto{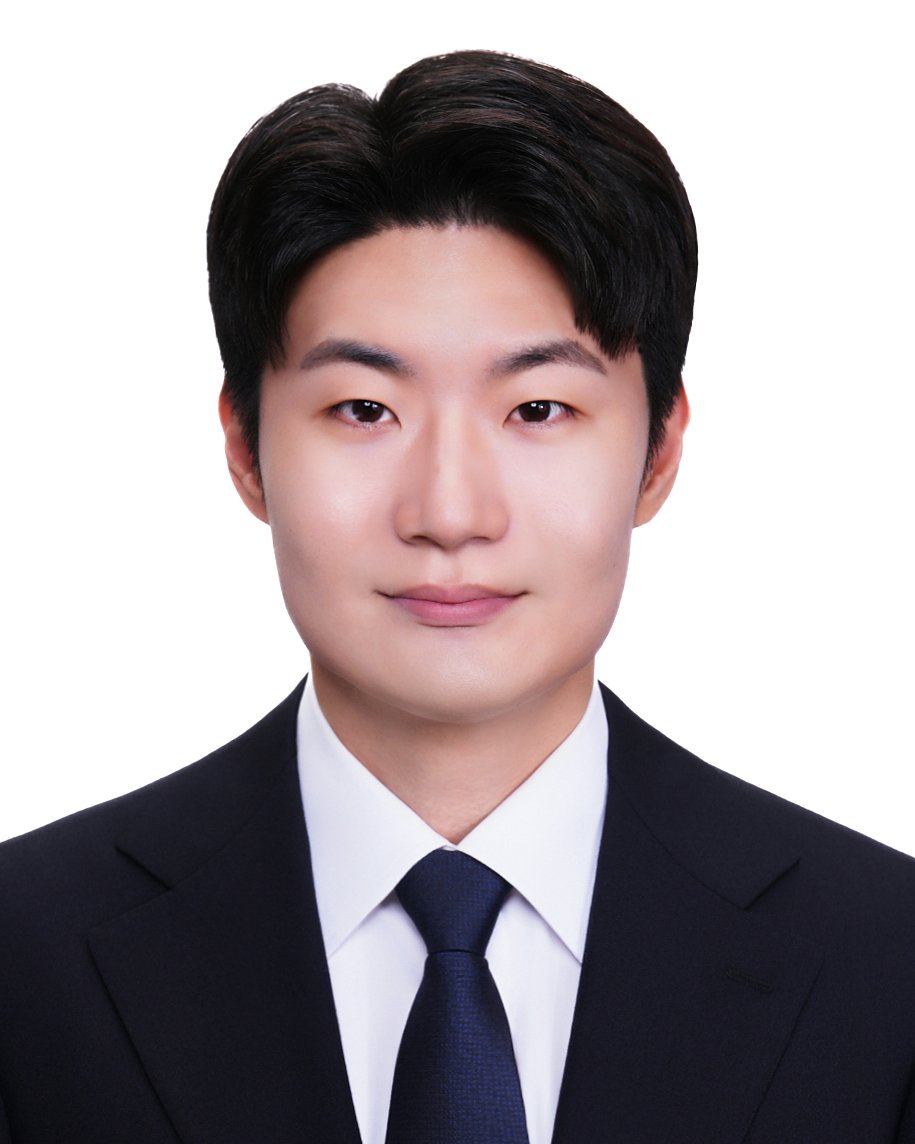}}]
{Kyung-Sik Choi}
(Member, IEEE) received the B.S., M.S., and Ph.D. degrees in electrical engineering from Korea Advanced Institute of Science and Technology (KAIST), Daejeon, South Korea, in 2016, 2018, and 2022, respectively.

From 2022 to 2024, he was a Post-Doctoral Researcher with ETH Zurich, Z\"urich, Switzerland. From 2024 to 2025, he was a tenure-track Assistant Professor with the Department of Electronic Engineering, Seoul National University of Science and Technology (SeoulTech), Seoul, South Korea. Since 2025, he has been a tenure-track Assistant Professor with the Department of System Semiconductor Engineering, Yonsei University, Seoul, South Korea. His research interests include radio frequency (RF) integrated circuits and systems, as well as high-frequency terahertz integrated circuits and systems based on complementary metal--oxide--semiconductor (CMOS) technology.
\end{IEEEbiography}

\end{document}